\documentclass{article}

\usepackage{microtype}
\usepackage{graphicx}
\usepackage{subfigure}
\usepackage{booktabs} 

\usepackage{pifont}%
\usepackage{comment}
\usepackage{amsmath}
\DeclareMathOperator*{\argmin}{argmin}

\usepackage{algorithm2e}
\usepackage{listings}
\usepackage[hyphens]{url}

\usepackage{hyperref}

\usepackage[accepted]{mlsys2025}

\mlsystitlerunning{Enabling Unstructured Sparse Acceleration on Structured Sparse Accelerators}

\begin{document}

\twocolumn[
\mlsystitle{Enabling Unstructured Sparse Acceleration on Structured Sparse Accelerators}



\mlsyssetsymbol{equal}{*}

\begin{mlsysauthorlist}
\mlsysauthor{Geonhwa Jeong}{to}
\mlsysauthor{Po-An Tsai}{nvr}
\mlsysauthor{Abhimanyu Bambhaniya}{to}
\mlsysauthor{Stephen W. Keckler}{nvr}
\mlsysauthor{Tushar Krishna}{to}
\end{mlsysauthorlist}

\mlsysaffiliation{to}{Georgia Institute of Technology}
\mlsysaffiliation{nvr}{NVIDIA}

\mlsyscorrespondingauthor{Geonhwa Jeong}{geonhwa.jeong@gatech.edu}

\mlsyskeywords{Machine Learning, MLSys}

\vskip 0.3in

\begin{abstract}

Exploiting sparsity in deep neural networks (DNNs) has been a promising area for meeting the growing computation requirements.
To minimize the overhead of sparse acceleration, hardware designers have proposed \emph{structured sparsity} support, but it provides limited flexibility and requires extra model fine-tuning. Moreover, any sparse model fine-tuned for certain structured sparse HW cannot be accelerated by other structured hardware.
To enable acceleration using unstructured sparsity of DNNs on structured sparse hardware, we propose an approximation method leveraging 
the distributive property in linear algebra to turn any sparse tensor into a series of structured sparse tensors.
We also develop a software framework, TASDER, to apply high-quality structured approximation on weights and activations of DNNs.
Our method accelerates dense and sparse DNNs without fine-tuning and improves energy-delay-product (EDP) by up to 83\% and 74\%. It achieves up to 39\% speed-up on a real system.
\end{abstract}

]



\printAffiliationsAndNotice{}  
\section{Introduction}


DNNs have revolutionized various domains~\cite{krizhevsky2012alexnet,convnext,vit, dlrm, gpt2,devlin-etal-2019-bert}, but DNN inference demands extreme compute power, memory footprints, and memory bandwidth as models scale to billions and trillions of parameters \cite{shoeybi2019megatron, fedus2021switch, chowdhery2022palm}.
To mitigate this growing demand, \emph{sparsity} has emerged as a promising opportunity. 
Model pruning~\cite{han2015deep} is the most popular method to remove a set of parameters in DNNs.
This optimization exploits a phenomenon that large models are often overly parameterized and do not need all parameters to maintain the target accuracy. 
Model pruning induces unstructured sparsity on weights (i.e. there is no specific pattern in the distribution of zeros) \text{unless} constraints are enforced during the pruning.
Also, many DNN models naturally exhibit activation sparsity due to the rectified linear unit (ReLU) that clips negative activation values to zeros resulting in unstructured sparsity.


Unfortunately, unstructured sparsity in DNNs leads to irregular memory accesses and diverged control/execution patterns, which are hostile to parallel architectures like GPUs, tensor cores~\cite{nvidia_volta}, and systolic arrays~\cite{jouppi2017tpu}. 
This has led to a plethora of academic research on unstructured sparsity support in DNN accelerators~\cite{parashar2017scnn, extensor,qin2020sigma,wang:dual-side, eureka2023micro}. 
Nonetheless, the complexity in hardware required for unstructured sparsity, such as indexing logic~\cite{extensor} and flexible distribution/reduction logic~\cite{qin2020sigma}, has been prohibitive in their commercial adoption.

Recent sparse DNN accelerators \cite{zhu2019micro,nvidia_ampere, liu2021s2ta, jeong:vegeta, AMD_MI300, Moffett_antoum} have proposed \textit{structured sparsity} support. 
The most popular example is \emph{fine-grained} N:M sparsity, 
which constrains at most N non-zeros in each block composed of M consecutive elements.
A commercial example of this is NVIDIA's sparse tensor core (STC) with 2:4 structured sparsity.
Structured sparse hardware forces DNN model developers to induce sparsity with certain constraints. \textit{Otherwise, that hardware is unusable.}
For example, NVIDIA’s fine-grained 2:4 structured sparsity support could provide 2$\times$ throughput by skipping ineffectual computations \textit{only if} the model is tuned with 2:4 structured sparsity. 
A model that does not follow 2:4 structured sparsity would, in fact, need to run on the dense tensor core, losing the benefits of sparsification.
On the flip side, a 2:4 structured sparse model cannot be accelerated on different structured sparse hardware, such as a sparse tensor core with 1:4 or 2:8 structured sparsity support.
Therefore, while the availability of structured sparse hardware is extremely promising, it remains limiting for model developers as it requires tuning the model to the specific sparsity support of the hardware, in addition to navigating the already vast design space, including DNN architectures and training recipes. This makes the broad adoption of structured sparse hardware challenging.
\begin{figure}[!t]
	\centering
	\includegraphics[width=0.49\textwidth]{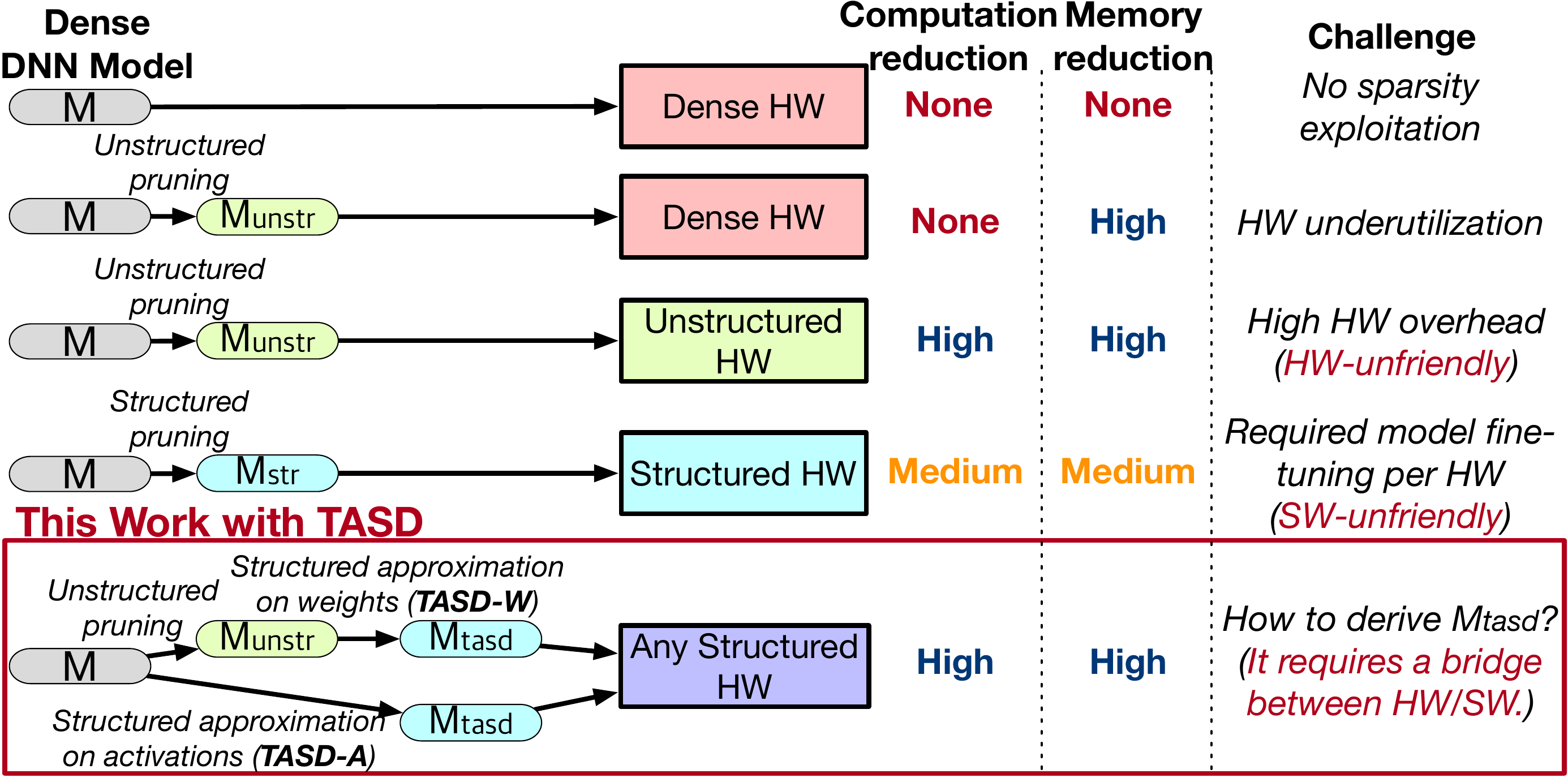}
 \vspace{-1em}
	\caption{Different flows to exploit sparsity in DNNs.}
 \vspace{-1em}
	\label{fig:tasd-high-level-fig}
\end{figure}

In this paper, we ask and answer the following  question \textbf{``Can we decouple the dependency between DNN models and HW, so that a model with \textit{any sparsity pattern} can be accelerated on \textit{any structured sparse HW}?}"
Our answer is to introduce a new level of abstraction between DL model developers and hardware designers. 
Specifically, we introduce a method, structured sparse tensor decomposition, to approximate any sparse tensor (such as pruned weights or intrinsic sparse activations) with a series of structured sparse tensors.
Leveraging the distributed property of tensor algebra,
we further propose to dynamically ``decode" an unstructured sparse tensor algebra into a series of 
structured sparse tensor algebra, which is efficient and compatible with prior structured sparse hardware. 
We contrast our proposed approach with current flows to exploit sparsity in \autoref{fig:tasd-high-level-fig}.

We make the following contributions: 
\begin{itemize}
    \item We propose Tensor Approximation via Structured Decomposition (TASD), which approximates any sparse tensor with a series of structured sparse tensors. TASD is the first work to demonstrate running unstructured sparse DNNs on structured sparse HW.
    \item We propose a framework, TASDER, a bridge between HW and SW to find the TASD series 
    to accelerate dense/sparse DNNs on structured sparse hardware.
    \item We propose a simple architectural extension and dataflow on top of existing structured sparse accelerators~\cite{jeong:vegeta} to execute TASD series efficiently.
    \item For various dense and sparse DNNs, we show that TASD improves EDP by up to 83\% and by 70\% on average. 
    We also show that across a range of DNNs, TASD can reduce the computation by 40\%.
    \item We show the effectiveness of TASD on commercial hardware (NVIDIA RTX 3080 GPU) with 2:4 sparse tensor cores and achieve up to 39\% performance gain.
\end{itemize}
\section{Background}

\subsection{Terminology}

\textbf{Sparsity} is a characteristic of data that includes zeros. 
The sparsity degree of a given tensor is measured as the fraction of the number of zeros to the number of the total elements in the tensor.
If a tensor has 0\% sparsity degree, we call the tensor \emph{dense}. 
Sparsity by itself is often used to indicate sparsity degree.
To describe the distribution of zeros, a \textit{sparsity pattern} can be given to the tensor.
If there is no defined sparsity pattern, we call it \emph{unstructured sparsity}.
Various patterns can be classified as structured sparsity, such as block sparse~\cite{narang2017block}, butterfly sparse~\cite{dao2021pixelated}, and mixed patterns~\cite{zaheer2020big}.

\begin{figure}[!t]
	\centering
	\includegraphics[width=0.46\textwidth]{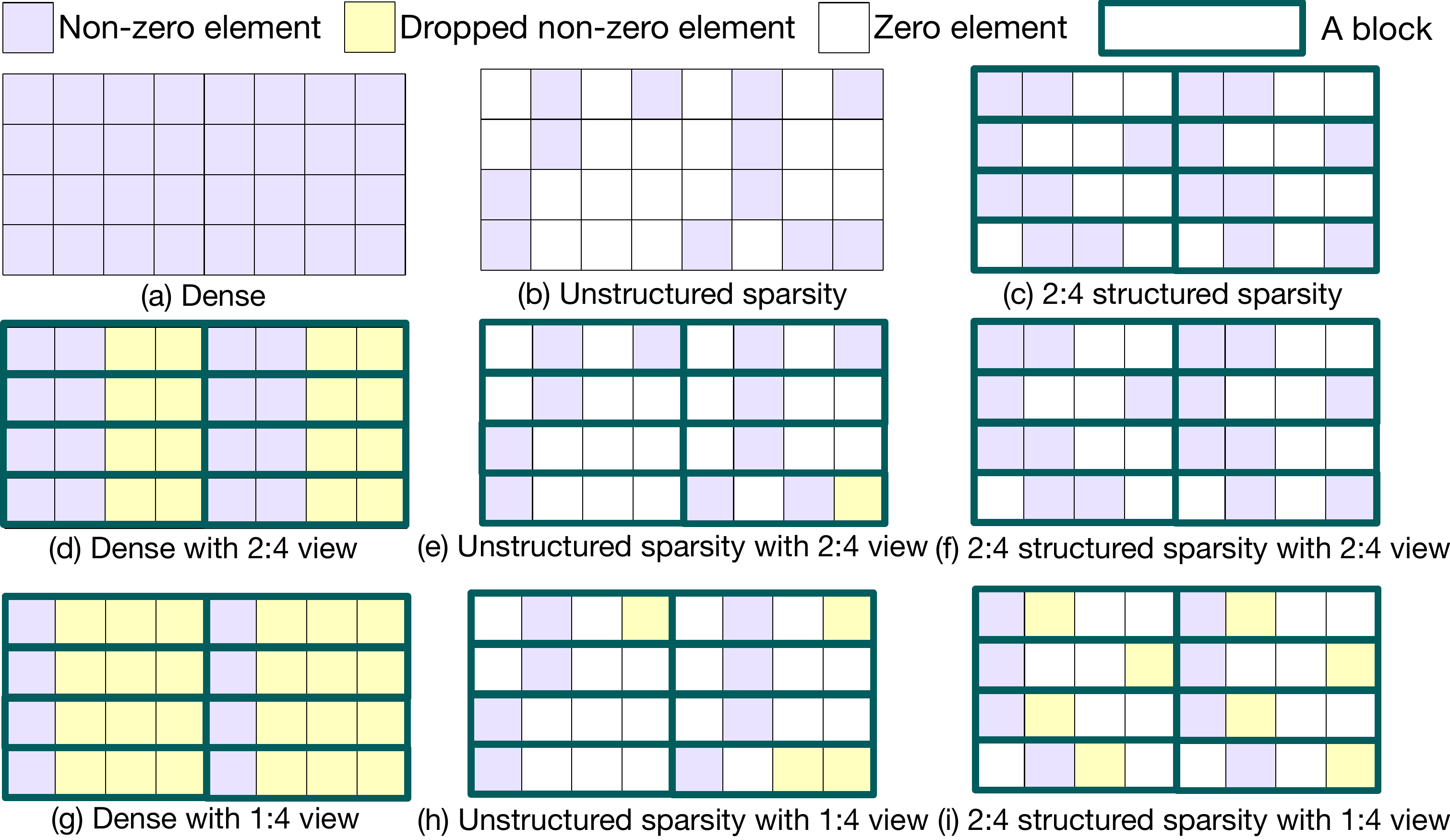}
 \vspace{-0.5em}
	\caption{Different sparsity patterns and views.
 }
 \vspace{-1em}
	\label{fig:sparsity_patterns}
\end{figure}

One of the most popular patterns is \textbf{N:M structured sparsity}~\cite{zhou2021nm}, as it is supported in both commercial products~\cite{nvidia_ampere, AMD_MI300, Moffett_antoum} and academic proposals~\cite{liu2021s2ta,jeong:vegeta, bambhaniya2023accelerating} with active training recipe research~\cite{pool2021channel,mishra2021accelerating, nvidia_asp, NM_training_recipe,lu2023step, sparsegpt, bambhaniya2024progressive}.
An N:M structured sparse tensor means that there can be at most N non-zeros in each M-element block in a certain rank of the tensor as shown in \autoref{fig:sparsity_patterns} (c).

We define \textbf{a view} of a tensor $A$ for a sparsity pattern as a tensor after potentially dropping some elements to meet the rule of the sparsity pattern.
For a matrix filled with non-zeros randomly, it is possible that the matrix does not meet the 2:4 sparsity pattern, i.e. there could be a block composed of 4 consecutive elements with more than two non-zeros.
To generate a 2:4 view of the matrix, some non-zeros in the matrix should be dropped (pruned in DNNs) to meet the pattern. As this process could drop some original values, it can be \textit{lossy}.
\autoref{fig:sparsity_patterns} also shows various tensors under different structured sparse views.

Prior work in DNN accelerators also proposed dense accelerators, unstructured sparse accelerators, 2:4 structured sparse accelerators, etc.
To clarify the nomenclatures, in this paper, \emph{if a sparsity pattern is used to describe a hardware accelerator, that accelerator should provide \textbf{lossless} and native support for any input tensor under such view.}
For example, we call Google TPU a dense accelerator and NVIDIA Sparse Tensor Core a 2:4 accelerator.

\subsection{DNN SW: Inducing sparsity in DNNs}
\textbf{Weight Sparsity.}
Without a specific pattern in mind, common model pruning methods introduces unstructured sparsity in weights~\cite{SparseZoo}. 
Due to the irregular accesses to handle non-zeros in unstructured sparse matrices, unstructured sparsity is not adequate for accelerating DNNs on the existing parallel hardware, such as GPUs.  

To address the issue, structured pruning forces pre-defined sparsity constraints in weights. For example, \textbf{N:M} structured sparsity~\cite{zhou2021nm,hubara2021accelerated,sun2021domino,mishra2021accelerating,fang2022algorithm,zhang2022learning,NM_training_recipe}
ensures there are at most \textbf{N} non-zeros in each block composed of \textbf{M} consecutive elements, so the required acceleration hardware can be trivial by exploiting the regularity in sparsity patterns.
Thus, various accelerators, including recent sparse tensor cores in NVIDIA Ampere GPUs~\cite{nvidia_ampere}, target to exploit the fine-grained structured sparsity instead of unstructured sparsity.
Nonetheless, structured pruning suffers from a higher loss of accuracy~\cite{sparsegpt} than unstructured pruning since the extra pruning constraints reduce the flexibility.
This extra loss of accuracy often leads to longer fine-tuning time (e.g., repeat the whole training process again) than unstructured sparse method to recover the loss in accuracy due to pruning~\cite{mishra2021accelerating}.



\textbf{Activation Sparsity.}
Activation sparsity arises at runtime due to the non-linear activation functions such as ReLU, ReLU6~\cite{mobilenets}, and SquaredReLU~\cite{so2021primer}, which clips negative values to zero. Activation sparsity is prevalent in both conventional Convolutional Neural Networks (CNNs) and recent Transformers~\cite{li:transformer-sparsity}. Since it is intrinsic in DNN models, no extra fine-tuning or pruning steps is required to introduce activation sparsity.
Unlike weight sparsity, activation sparsity is dynamic as the intermediate input activation values depend on the inputs of the DNN model.
Thus, the location of non-zeros and the degree of sparsity are unpredictable, similar to unstructured pruning making it hard for structured sparse hardware to exploit input sparsity.
Another challenge is that recently proposed activation functions, such as GELU~\cite{gelu}, and Swish~\cite{ramachandran2017searching}, do not generate zero, which nullifies the benefits of exploiting activation sparsity in prior work~\cite{jang2021samsung}.

\subsection{DNN HW: Exploiting sparsity in DNNs}
\label{subsec:dnnhw}


Unstructured sparse accelerators, including SCNN \cite{parashar2017scnn}, SIGMA \cite{qin2020sigma}, Samsung NPU~\cite{jang2021samsung}, and dual-side sparse core (DSTC)~\cite{wang:dual-side} target unstructured sparsity and skip redundant computations aggressively, but they suffer from non-trivial area/power costs due to the complex indexing and reduction logic, often introducing workload imbalance problems~\cite{liu2022hpca} as well. 
For example, SIGMA \cite{qin2020sigma} introduces 38\% area overhead compared to the dense architecture due to its flexible and non-blocking distribution/reduction networks.
SCNN \cite{parashar2017scnn} and Griffin~\cite{shin2021griffin} produce 34\% and 32\% area overhead due to the support for the unstructured sparse dataflow.
Moreover, when the sparsity degree is low or zero, they provide no improvement or even degrade performance/efficiency due to the overhead for supporting unstructured sparsity \cite{highlight2023micro}.
More recent structured sparse tensor accelerator architectures, such as STA \cite{liu2020sta}, Sparse Tensor Core from NVIDIA GPUs (NV-STC) \cite{nvidia_ampere}, and VEGETA \cite{jeong:vegeta}, provide HW support for structured sparsity
with minimal area overhead.
However, these designs accelerate \textbf{only structured pruned models} with the specific pattern and focus on weight sparsity since exploiting unstructured activation sparsity without much overhead is not trivial.
S2TA~\cite{liu2022hpca} has tried to circumvent the challenge by forcing structured sparse patterns dynamically, but it requires modifying the models and even more fine-tuning steps. 
We compare different DNN HWs in \autoref{table:nm-hw}.
\definecolor{darkred}{rgb}{.65,0,0}
\definecolor{darkgreen}{rgb}{0,.5,0}
\definecolor{darkyellow}{rgb}{0.95,.6,0.1}
\newcommand{\xmark}{ \textcolor{darkred}{\ding{56}}}
\newcommand{\hmark}{ 
{\textbf{--}}}
\newcommand{\upmark}{ \textcolor{darkgreen}{\ding{115}}}
\newcommand{\downmark}{ \textcolor{darkred}{\ding{116}}}

\newcommand{\cmark}{\textcolor{darkgreen}{\ding{51}}}%
\begin{small}
\begin{table*}[!ht] 
\small
\centering
\caption{Comparison of DNN HWs. Unstr: Unstructured sparse. Str: Structured sparse. Wgt: Weights. Act: Activations.}
\begin{tabular}{|c|c|c|c|c|c||c|}
\hline
HW Support
&
{\textbf{Dense}}
&
{\textbf{Unstr}}
&
{\textbf{Str}}
&
{\textbf{Dense}}
&
{\textbf{Unstr}}
&
{\textbf{Area}}

\\
$\downarrow$
&
{\textbf{Wgt}}
&
{\textbf{Wgt}}
&
{\textbf{Wgt}}
&
{\textbf{Act}}
&{\textbf{Act}}
&
{\textbf{Cost}}

\\

\hline
\hline

\begin{tabular}[c]{@{}c@{}}
Dense~\cite{jouppi2017tpu, nvidia_volta, rasa}
\end{tabular}
& \cmark
& \xmark
& \xmark 
& \cmark
& \xmark
& \cmark\cmark
\\
\hline

\begin{tabular}[c]{@{}c@{}}
Unstr~\cite{qin2020sigma,parashar2017scnn,wang:dual-side}
\end{tabular}
& \xmark\renewcommand{\thefootnote}{\fnsymbol{footnote}}\footnotemark[1]
& \cmark
& \cmark 
& \xmark\renewcommand{\thefootnote}{\fnsymbol{footnote}}\footnotemark[1]
& \cmark
& \xmark
\\
\hline

\begin{tabular}[c]{@{}c@{}}
Str~\cite{zhu2019micro, liu2020sta, nvidia_ampere,jeong:vegeta}
\end{tabular}
& \cmark
& \xmark
& \cmark
& \cmark
& \xmark
& \cmark
\\
\hline

\begin{tabular}[c]{@{}c@{}}
\textbf{TASD} (This work)
\end{tabular}
& \cmark 
& \cmark
& \cmark
& \cmark\cmark\renewcommand{\thefootnote}{\fnsymbol{footnote}}\footnotemark[7]
& \cmark 
& \cmark
\\
\hline

\end{tabular}

\hfill
\renewcommand{\thefootnote}{\fnsymbol{footnote}}
\begin{minipage}{1.0\linewidth}
\footnotemark[1]{With extra wiring/logic, unstructured sparse HW is inefficient if the tensor is dense.}\mbox{}\hfill
\end{minipage}
\begin{minipage}{1.0\linewidth}
\footnotemark[7]{TASD enables further acceleration by approximating dense tensors with sparse tensors.}\mbox{}\hfill
\end{minipage}

\label{table:nm-hw}
\end{table*}
\end{small}



\begin{figure}[!t]
	\centering
	\includegraphics[width=0.49\textwidth]{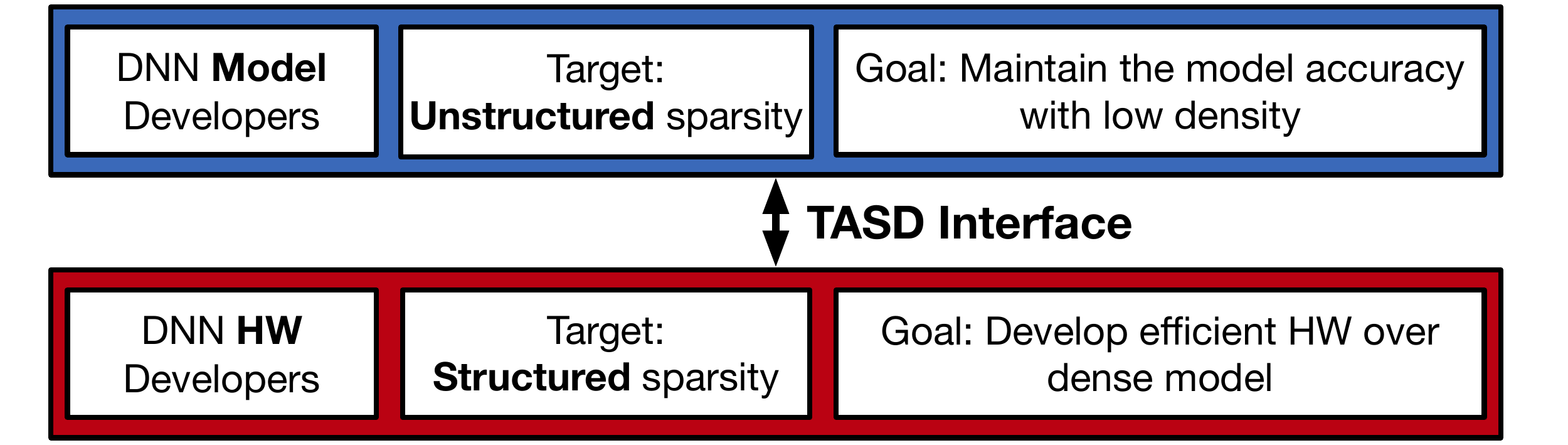}
        \vspace{-1.5em}
	\caption{TASD Interface.
 }
        \vspace{-1em}
	\label{fig:tasd_inteface}
\end{figure}
\subsection{Tension between sparse DNN SW and HW}
\autoref{fig:tasd_inteface} shows the state of sparse DNN software and hardware. On the one hand, model developers have shown that unstructured sparsity provides better model accuracy and higher sparsity degree. On the other hand, hardware developers have shown that structured sparsity support is more practical to include in GPUs and other DNN accelerators. Such tension in the desired sparsity patterns hampers the progress in bringing sparse DNN acceleration to practice.

The main drawback of the previous hardware-specific patterns is that the pruning software and hardware support are tightly coupled, such that the software generates a model specifically pruned for the pattern supported by the hardware. For example, a model pruned for the NV-STC can only be accelerated by NV-STC, not by S2TA.
To decouple the tight relationship, we propose another layer of system software between the model developers and DNN hardware for sparsity. Our insight is to approximate a tensor by decomposing it into a series of structured sparse tensors. 
We leverage the distributive property in tensor algebra to execute the series of structured sparse GEMM. This mechanism provides an unstructured sparse interface for developers but only requires structured sparse support from hardware. As shown in \autoref{table:nm-hw}, by bridging DNN model and HW, this work is able to accelerate all types of sparsity seamlessly with a low area overhead.


\section{TASD: Tensor Approximation via Structured Decomposition} 

In this section, we introduce a method to \textbf{approximate unstructured sparsity using a series of structured sparsity}, which we call \textbf{TASD}.
In this paper, we use a set of N:M structured sparsities for TASD to explain the method and show how to use it practically, but the method is general and not limited to only N:M structured sparsity.

\label{sec:tasd}
\begin{figure}[!t]
	\centering
	\includegraphics[width=0.49\textwidth]{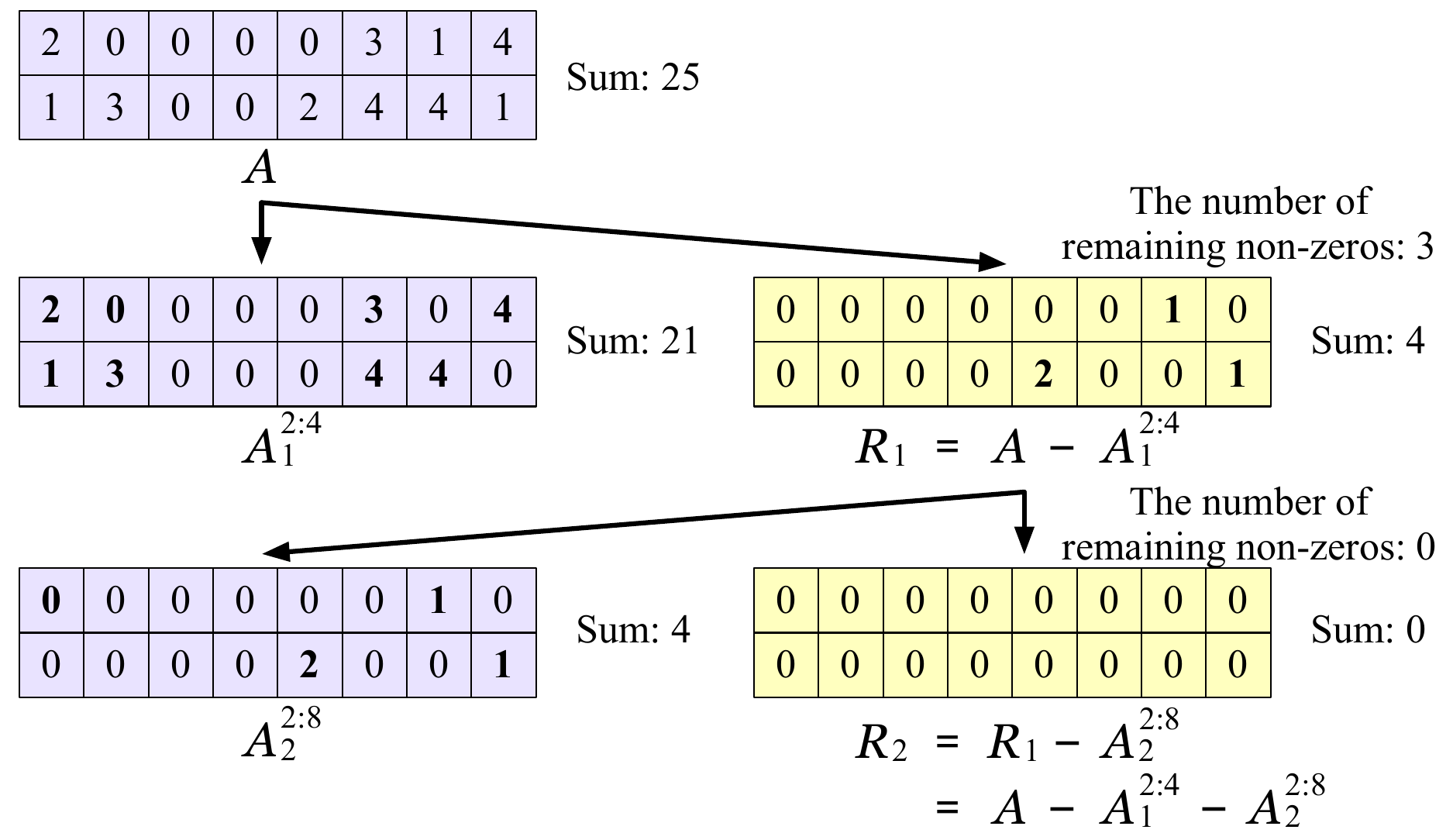}
        \vspace{-1em}
        \caption{TASD example using a 2$\times$8 matrix $A$.}
	\label{fig:tasd_example}
    \vspace{-2em}
\end{figure}

\subsection{Overview}
We use an unstructured sparse matrix $A$ to illustrate how TASD works in \autoref{fig:tasd_example}.
The matrix $A$ has 6 zero elements out of 16 total elements with a 37.5\% sparsity degree. 
Also, note that the sum of all elements in $A$ is 25.

Matrix $A$ can be rewritten as a 2:4 structured sparse matrix (a 2:4 view of $A$) plus a remaining matrix, $A_{1}^{2:4}$ and $R_1$, where $A_{1}^{2:4}$ is derived by extracting two \textbf{largest} elements out of four elements in each row in $A$ while $R_1$ is the remaining matrix (i.e. $A - A_{1}^{2:4}$) after the extraction, as shown in \autoref{ss-decomp}. 
\begin{align}
    A & = A_{1}^{2:4} + R_1 \label{ss-decomp}
\end{align}
The extracted matrix, $A_{1}^{2:4}$ covers 70\% in terms of the number of non-zero values while covering 84\% in terms of the sum of the magnitudes.
The percentage for the lost magnitudes is smaller than the percentage of the lost non-zero values because we extract two \textit{largest} elements out of four consecutive elements.
If we discard the remaining matrix $R_1$, then the original matrix $A$ can be \textit{approximated} as $A_{1}^{2:4}$.
Thus, we call this approximation, \textbf{structured decomposition}.
If we approximate $A$ with a 3:4 pattern instead of the 2:4 pattern, we can derive matrix $A_{1}^{3:4}$ with a structured decomposition that drops only one non-zero element, covering 90\% of the number non-zeros and 96\% of the sum of total magnitudes.

Instead of using a denser N:M, we can further decompose $R_1$ using another structured pattern, such as 2:8.
$A_2$ can also be derived by extracting two largest elements out of \textbf{eight} consecutive elements in $R_1$, making $A_2$ as a 2:8 structured sparse matrix.
Similar to the previous decomposition, we call the remaining matrix $R_2$ as shown in \autoref{fig:tasd_example}.
All elements of $A$ are covered by $A_{1}^{2:4}$ and $A_{2}^{2:8}$, so $A$ is equal to $A_{1}^{2:4} + A_{2}^{2:8}$, thus the approximation of $A$ to $A_{1}^{2:4} + A_{2}^{2:8}$ is lossless.
Since the unstructured sparse matrix is approximated using a set of structured sparse matrices, we call this method as Tensor Approximation via Structured Decomposition (TASD).

Theoretically, structured decomposition can have infinite terms. 
Below, we formalize the process more generally using different structured sparse patterns denoted as $s_i$.
\begin{align}
    A & \simeq A_1^{s_1} + A_2^{s_2} \label{ss-decomp2}\\ \label{ss-decomp3}
      & \simeq A_1^{s_1} + A_2^{s_2} + A_3^{s_3} \\ \label{ss-decomp4}
      & \simeq A_1^{s_1} + A_2^{s_2} + A_3^{s_3} ... + A_n^{s_n}
\end{align}
We call $A$ as the original matrix and $\sum A_i^{s_i}$ as \textbf{TASD series}, to draw an analogy to the classic Taylor series%
\footnote{Taylor series approximates any function with polynomials, while TASD series approximates any tensor with structured sparse tensors.}%
: Each successive term (residual structure sparse tensor) improves the accuracy of the approximation.
A TASD series configuration includes the number (``order") of TASD terms ($n$) and the structured sparsity pattern ($s_i$) for each TASD term.
Using TASD, one can generate a structured sparse view of a given tensor, and the error between the view and the original tensor would depend on the TASD series configuration.

\subsection{Using TASD for matrix multiplication}
TASD decomposes any tensor $A$ into a series of structured sparse tensors.
Decomposed tensors can be used in any tensor algebra, such as matrix multiplication ($C = A\times B$), which can be approximated as $A_1^{s_1}\times B$. If $s_1$ is 2:4, and the matrix multiplication is running on NVIDIA Sparse Tensor Cores, potentially 50\% of  Multiply-and-Accumulate (MAC) operations could be skipped.

With the distributive property of tensor algebra, matrix $A$ can be approximated using more TASD terms such as $(A_1^{s_1} + A_2^{s_2})B = A_1^{s_1}B + A_2^{s_2}B$. 
If $s_1$ is 2:4 and $s_2$ is 2:8, about 25\% of MAC operations could be skipped.
Finding the proper TASD series to minimize the error while maximizing compute reduction will determine the quality of the approximation.
We provide a detailed analysis of TASD with synthetic data in \autoref{append:additional_details}.

\section{HW/SW Co-Design with TASD}
\label{sec:codesgin}

\begin{figure}[!t]
	\centering
	\includegraphics[width=0.49\textwidth]{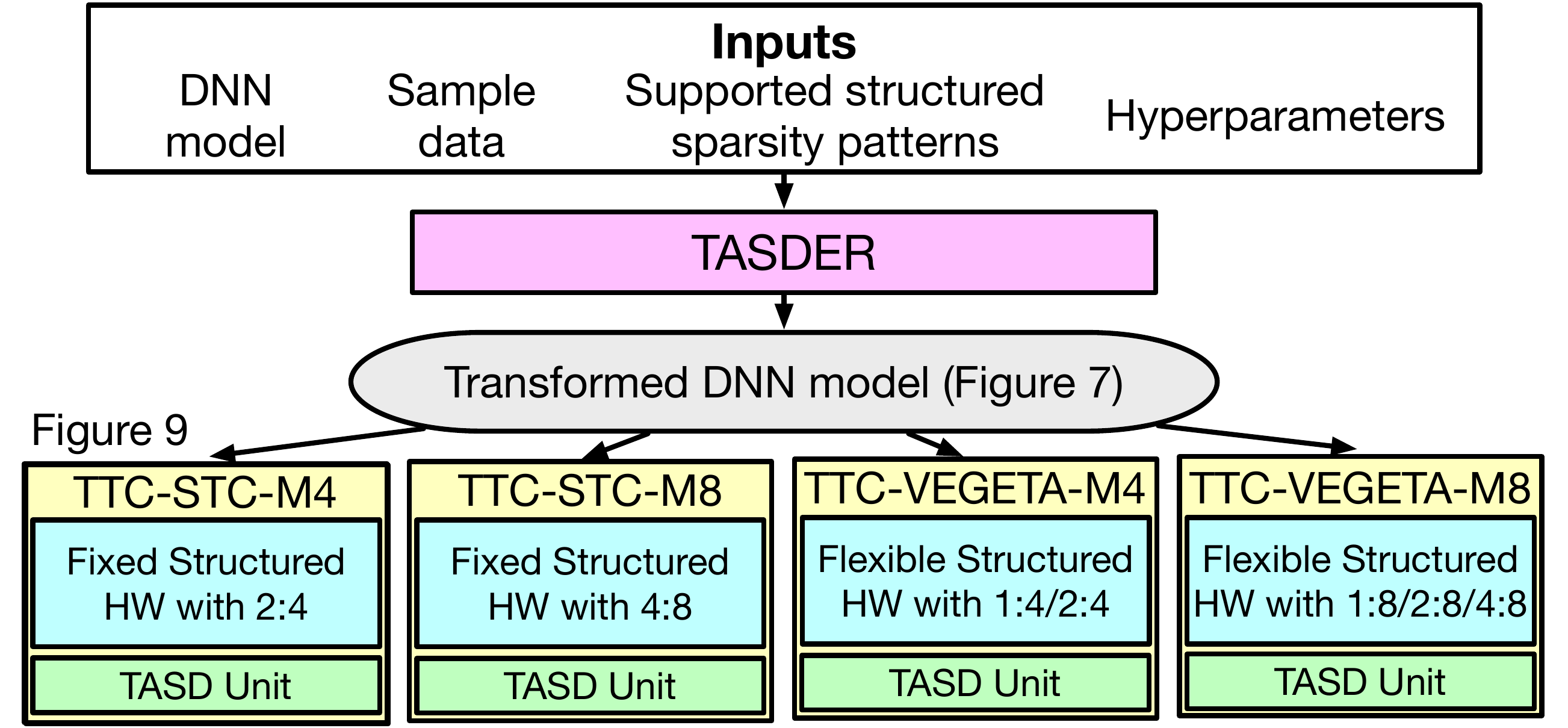}
    \vspace{-0.5em}
	\caption{System overview with TASDER.}
	\label{fig:system_overview}
    \vspace{-0.5em}
\end{figure}

In \autoref{sec:tasd}, we introduce our approximation method, TASD, in general. In this section, we show how our method can be used to accelerate DNN models with sparse weights and inputs.
Although TASD can also be used to accelerate the DNN training, we focus on how to accelerate DNN inference in this work.
There are two main insights that inspired us to use TASD for DNN inference.

1) By its nature, DNN models are able to tolerate small errors in their internal computations. 

2) Although TASD is a lossy approximation method, carefully selected TASD terms can provide high-quality approximations with a limited number of non-zeros being dropped.


\subsection{System architecture overview}
We introduce our optimizer system, TASDER, which takes a DNN model, sample data, target HW information including structured sparsity patterns, and hyperparameters as shown in \autoref{fig:system_overview}.
Internally, TASDER will search for the TASD configuration for each layer of the given DNN model and return the configurations.
In the following subsections, we introduce TASD-W and TASD-A, which are the methods to exploit TASD on weights and activations, respectively.
We also explain how the TASD configuration per layer is selected in our framework.
In this work, we only consider convolution (CONV) and fully-connected (FC) layers in DNN models to apply TASD as they usually consume most of the execution cycles, and they get converted to matrix multiplication operations using algorithms such as \texttt{im2col} for parallelization.


\subsection{TASD-W: Applying TASD on weights}
\label{subsec:tasd-w}
\begin{figure}[!t]
	\centering
	\includegraphics[width=0.45\textwidth]{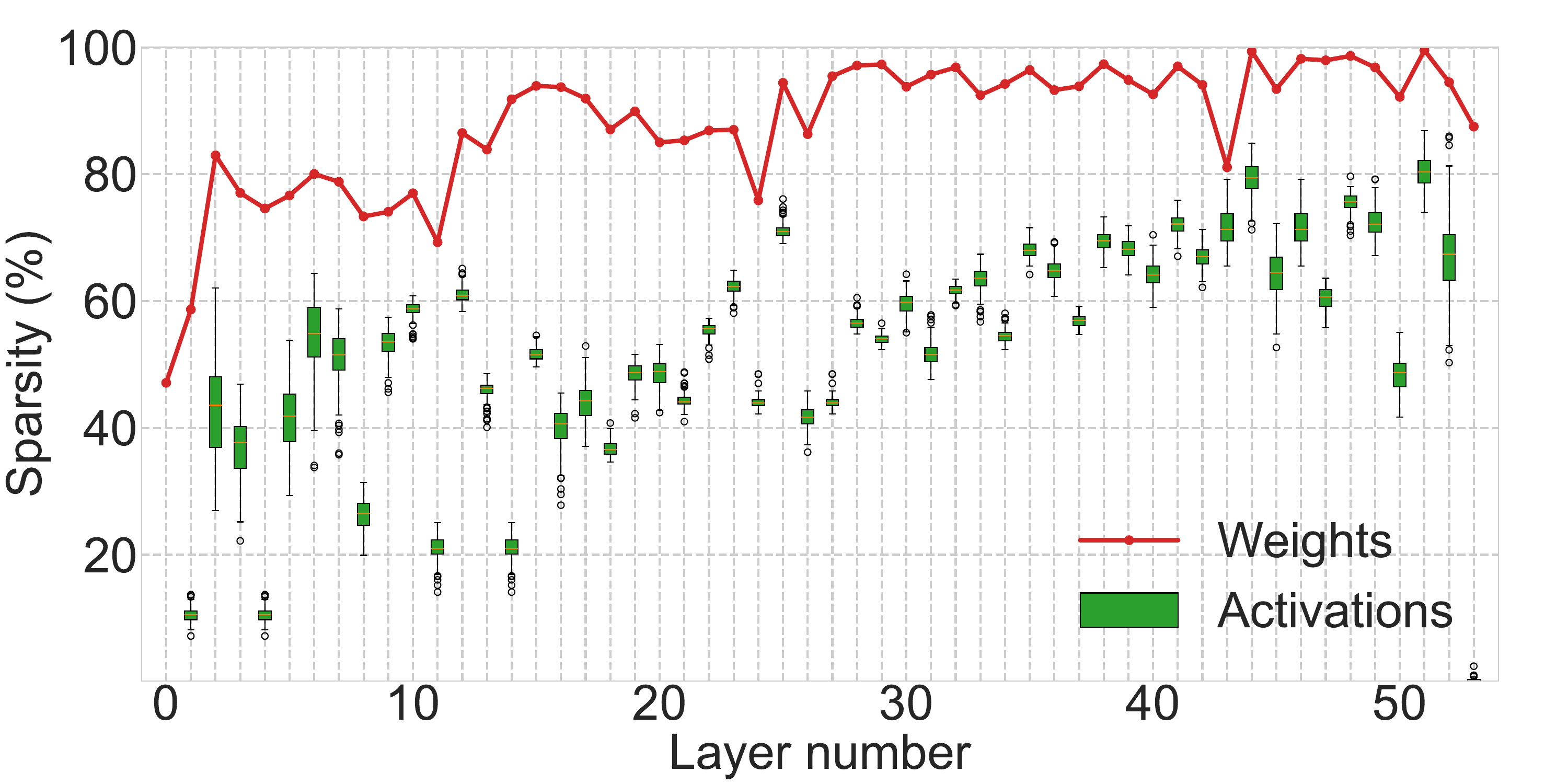}
        \vspace{-1em}
	\caption{Sparsity degrees for each layer in 95\% unstructured sparse ResNet50 from SparseZoo.}
	\label{fig:resnet50_layer_sparsity}
    \vspace{-1em}
\end{figure}

\begin{figure*}[!t]
	\centering
	\includegraphics[width=0.95\textwidth]{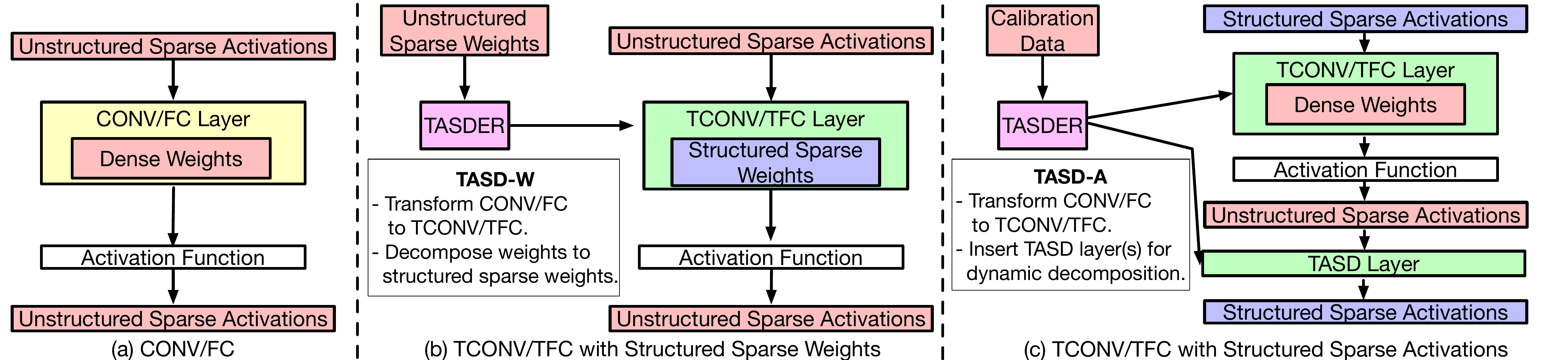}
        \vspace{-0em}
	\caption{Comparison of flows of an original model with a conventional CONV/FC layer and transformed models with a TCONV/TFC layer with structured sparse weights, and a TCONV/TFC/TASD layer with structured sparse activations. 
 }
	\label{fig:conv-tconv}
    \vspace{-1em}
\end{figure*}

We expose an unstructured interface to ML model developers as the target of optimization so that they can focus on the techniques to prune their models as much as possible without considering any specific HW-friendly sparsity pattern.
Therefore, the optimization problem for TASD-W is that given the weights of DNN models as unstructured sparse tensors, use the available structured sparse HW to accelerate the model execution as much as possible. 

We assume that the target hardware can accelerate the structured sparsity patterns, $S_1, ..., S_n$.
A TASD configuration of the $i$th layer, $C_i$, is a sequence of $S$.
For a given DNN model $M$, a TASD transformation of the model, $T$, is defined as applying a sequence of $C_i$ where $C_i$ is the TASD configuration for each layer in the model.
Then, the target is to find a TASD transformation $T_{opt}$ for a given model where
\vspace{-0.5em}
\begin{align}
& T_{opt} = \argmin_T (Latency(M_T)) \\  
& \text{such that } Accuracy(M_T) \approx {Accuracy}(M_{original}) 
\end{align}

\vspace{-0.5em}
A simple way to use TASD-W is using the same TASD configuration for all layers in  the model, i.e. applying \textbf{network-wise TASD-W}.
As the number of supported structured sparsity patterns, $n$, is not large enough, the $T_{opt}$ for network-wise TASD-W could be found with the exhaustive search.

A better method to use TASD-W is using different TASD configurations for different layers, i.e. \textbf{layer-wise TASD-W}.
The set of TASD transformations that can be covered by layer-wise TASD-W is a super-set of the TASD transformations in network-wise TASD-W. 
Usually, a pruned model with unstructured sparsity does not guarantee the same sparsity across layers, i.e. even though the overall sparsity is 95\% for the model, different layers could have different sparsity degrees as shown in \autoref{fig:resnet50_layer_sparsity}.
Unlike network-wise TASD-W where all the layers use the same TASD configuration, it is not straightforward how to choose a TASD configuration per layer as there could be numerous options per layer. 
To minimize the accuracy drop, it is crucial to reduce the number of dropped non-zeros after applying TASD, which would prefer conservative TASD configurations. 
On the other hand, to maximize the performance gain, it would be better to apply aggressive TASD configurations, which would be able to be translated into higher sparsity and efficiency gain. 

To address this, we design and implement a greedy-based algorithm that optimizes across all layers.
This greedy algorithm first measures the percentage of dropped non-zero elements of each TASD configuration for all layers and sorts the configuration-layer pairs by their percentage of dropped elements.
Next, it greedily applies the TASD configuration based on the sorted order until the model quality is $<$99\% of the original model (i.e., prioritize the option with the smallest dropped non-zeros). 
Since it only takes a single pass to all layers, the runtime overhead is trivial (a few seconds per model), while the extra training needed for structured sparse HW often takes hundreds of GPU hours~\cite{sun2021domino}.

We use TCONV/TFC to indicate a CONV/FC layer with TASD as shown in \autoref{fig:conv-tconv}, and the TASD configurations found above would be applied to the corresponding TCONV/TFC layers.
In \autoref{fig:conv-tconv} (a), we show how the conventional CONV/FC layer works with unstructured sparse activations (usually from ReLU) with dense weights. In \autoref{fig:conv-tconv} (b) shows how a TCONV/TFC layer works with unstructured sparse weight. 
TASDER would modify unstructured sparse weights to structured sparse weights with TASD-W.

\subsection{TASD-A: Applying TASD on activations}
\begin{figure}[!t]
	\centering
	\includegraphics[width=0.48\textwidth]{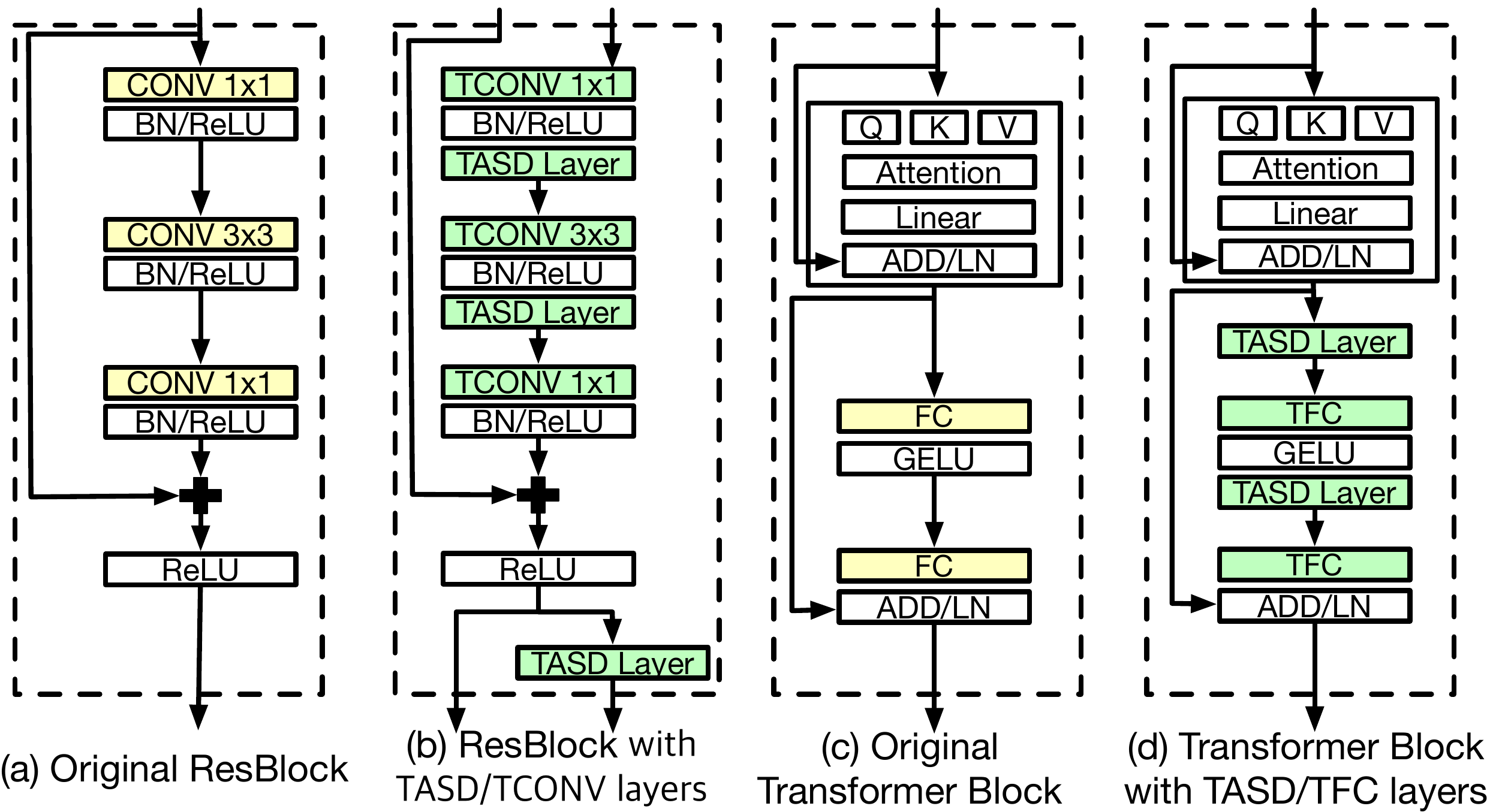}
 \vspace{-1.5em}
	\caption{
 ResBlocks with CONV and TASD/TCONV and Transformer Blocks with FC and TASD/FC.}
 
	\label{fig:tasd-a}
 \vspace{-1.5em}
\end{figure}
TASD can also be applied to activations to improve a DNN execution as shown in \autoref{fig:conv-tconv} (c).
Unlike weights which are static, TASD should be applied for dynamic decomposition during the runtime as activations are dynamic. 
In \autoref{fig:tasd-a} (a) and (b), we show a baseline ResBlock and a ResBlock with TCONV and TASD layers, which decompose activation tensors with given TASD configurations. 
We add TASD layers after ReLU layers so that they can minimize the number of dropped non-zeros after approximation. 
The same can be applied to a Transformer Block, where the FC layers in the multi-layer perception module can be replaced with TFC by inserting a TASD layer before TFC layers as shown in \autoref{fig:tasd-a} (c) and (d).
Ideally, other FC layers in a Transformer Block could also be replaced with TASD and TFC layers, but empirically we found it hard to maintain the model quality.

Similar to TASD-W, the simplest way to choose a TASD configuration for each TASD layer is using network-wise TASD where all TASD configurations are same across all TASD layers. 
Assuming limited supported structured sparsity patterns from HW, only a handful number of options need to be explored.
However, similar to weights, this may not be efficient as activations from different layers show significantly different degrees of sparsity, as shown in \autoref{fig:resnet50_layer_sparsity}.

To address this issue, we again leverage layer-wise TASD as it can tailor the TASD configuration to each layer.
However, 
unlike the TASD-W,
it is not feasible to test every option for each layer to find out the best options as the target tensors (activations) are dynamically generated.
We find that a small set of calibration dataset (e.g., 1000 images for ImageNet~\cite{deng2009imagenet}) can provide enough insights. 
As shown in \autoref{fig:resnet50_layer_sparsity}, while different layers have different sparsity degrees, for a particular layer, the activation sparsity degree remains in a small range across different input images.
Therefore, TASDER takes calibration data as an input, so it can profile the given DNN model with the calibration data and collect the statistics (e.g., average, 99th percentile) about activation sparsity per layer. 

To choose a TASD-A configuration for each layer, we use a \emph{sparsity-based} selection method, instead of the dropped-non-zero-based method for TASD-W.
We use a hyperparameter, $\alpha$, to tune the aggressiveness of the TASD approximation. 
For a given layer $L_i$ and the available configurations in the target HW (e.g., $H_1, ... , H_4$), we choose $C_i$ as $H_j$ where $j$ is the largest integer where $S(L_i) + \alpha > H_j$.
If we use a larger $\alpha$, we choose the TASD configuration more aggressively (i.e. allowing more dropped non-zeros). 


\textbf{Beyond sparsity: Supporting non-ReLU-based DNNs.}
For better accuracy, state-of-the-art DNNs have replaced ReLU with other activation functions, such as GeLU \cite{gelu} and Swish~\cite{ramachandran2017searching}, which do not induce any sparsity in activations making the activations dense.
Thus, our sparsity-degree-based TASD selection to choose TASD configuration for TASD-A for each layer would not work for those DNNs.

\begin{table}[t]
\small
\caption{Supported sparse patterns with TTC-VEGETA.}
\begin{center}
\begin{tabular}{cc|cc}
{\bf Pattern} & {\bf TASD series} & {\bf Pattern} & {\bf TASD series}\\
\hline
1:8 & 1:8 & 5:8 & 4:8 + 1:8\\
2:8 & 2:8 & 6:8 & 4:8 + 2:8\\
3:8 & 2:8 + 1:8 & 7:8 & - \\
4:8 & 4:8  & 8:8 & Dense\\

\end{tabular}
\label{tab-ttc-vegeta}
\end{center}
\vspace{-1em}
\end{table}

To address this, we investigate the distribution of the magnitude of all elements in the activation tensors from GeLU/Swish-based DNN.
We found that, while no element in the tensor is exactly zero, a huge number of elements have tiny magnitude, compared to the range of magnitude for all elements. Therefore, we let TASD-A leverage this skewed distribution and collect the magnitude statistics. We introduce another heuristic, \emph{pseudo-density}, which aims to preserve a fixed percentage (e.g., 99\%) of the sum of all elements in a tensor, to determine the best TASD configuration for every layer.
Using the pseudo-density for the non-ReLU-based DNNs, we can use the same sparsity-degree-based method 
(i.e. by replacing \textit{sparsity} to \textit{1 - pseudo-density}). The approximating nature of TASD allows the system to also accelerate non-ReLU-based DNNs, while prior work that specifically targets activation sparsity cannot.




\begin{figure}[!t]
	\centering
	\includegraphics[width=0.45\textwidth]{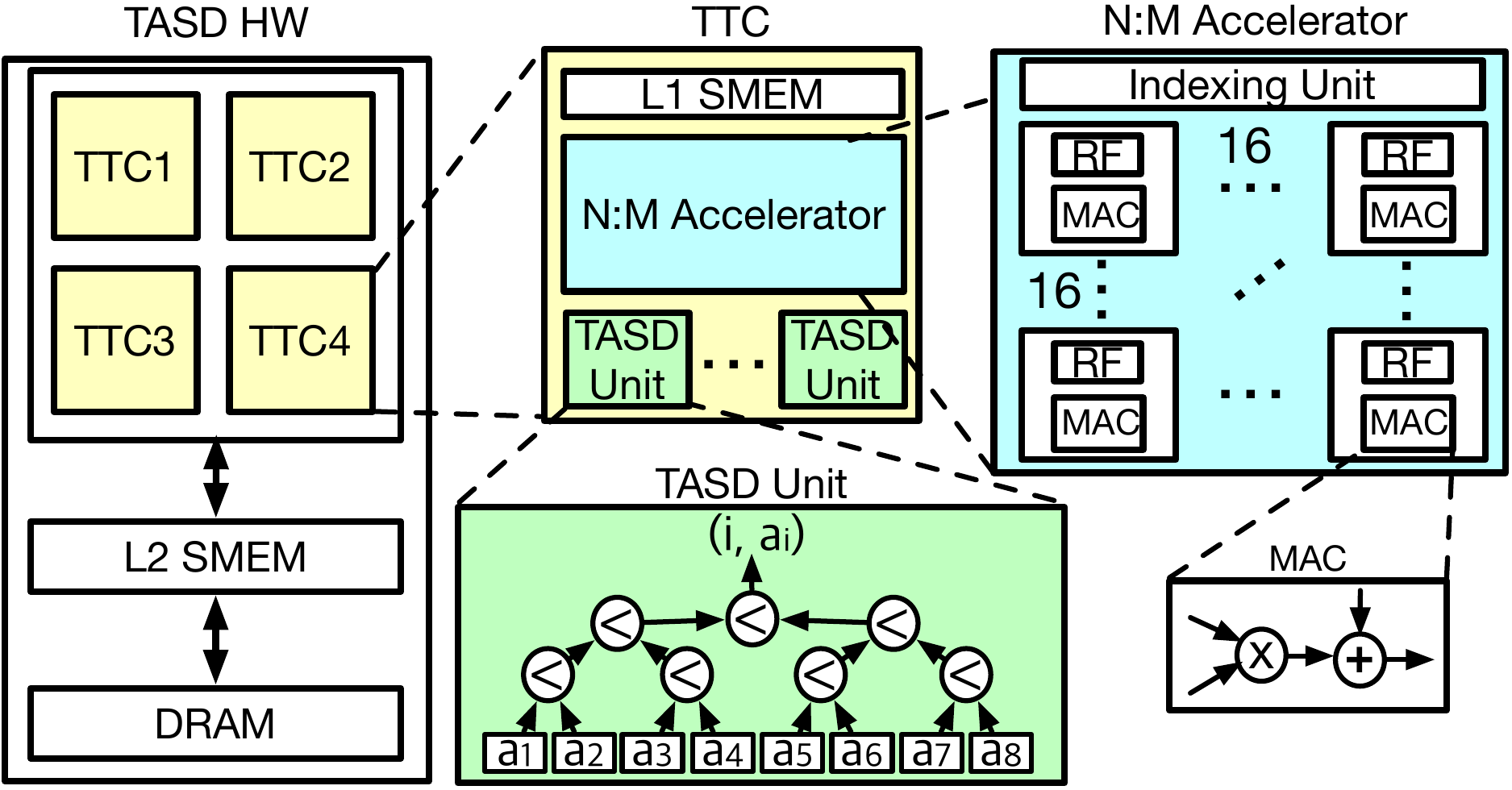}
  \vspace{-1em}
	\caption{TASD-HW composed of four TTCs. 
 }
	\label{fig:tasd-hw}
 \vspace{-1em}
\end{figure}
\subsection{Structured sparse HW for TASD}
TASD works best when there are at least a few supported structured sparse patterns in the target sparse accelerator. While the TASDER optimizer is HW-agnostic, we propose to build on top of a recently proposed flexible structured sparse tensor accelerator to maximize the benefit.
Inspired by previous structured sparse accelerators~\cite{nvidia_ampere,liu2021s2ta,jeong:vegeta}, we introduce TASD Tensor Core (TTC).
We adopt a design similar to VEGETA~\cite{jeong:vegeta} engine composed of multiple processing elements (PEs) while providing support for 1:8, 2:8, and 4:8 structured sparse patterns, and we call it TTC-VEGETA.
With TASD and a limit of up to 2 terms, a TTC-VEGETA engine can support 7
out of all the N:8 patterns 
as shown in \autoref{tab-ttc-vegeta} even though the original VEGETA supports only three sparse patterns.
Note that TTC can adopt other structured sparse designs, such as STC~\cite{zhu2019micro} with supports for 2:4 and dense, which we call TTC-STC. 
This would limit the flexibility in approximation using TASD compared to VEGETA-based TTC design, but TASDER is still able to optimize some layers. 
We explore the benefit of flexibility in \autoref{sec:eval}.

\begin{figure}[!t]
	\centering
	\includegraphics[width=0.45\textwidth]{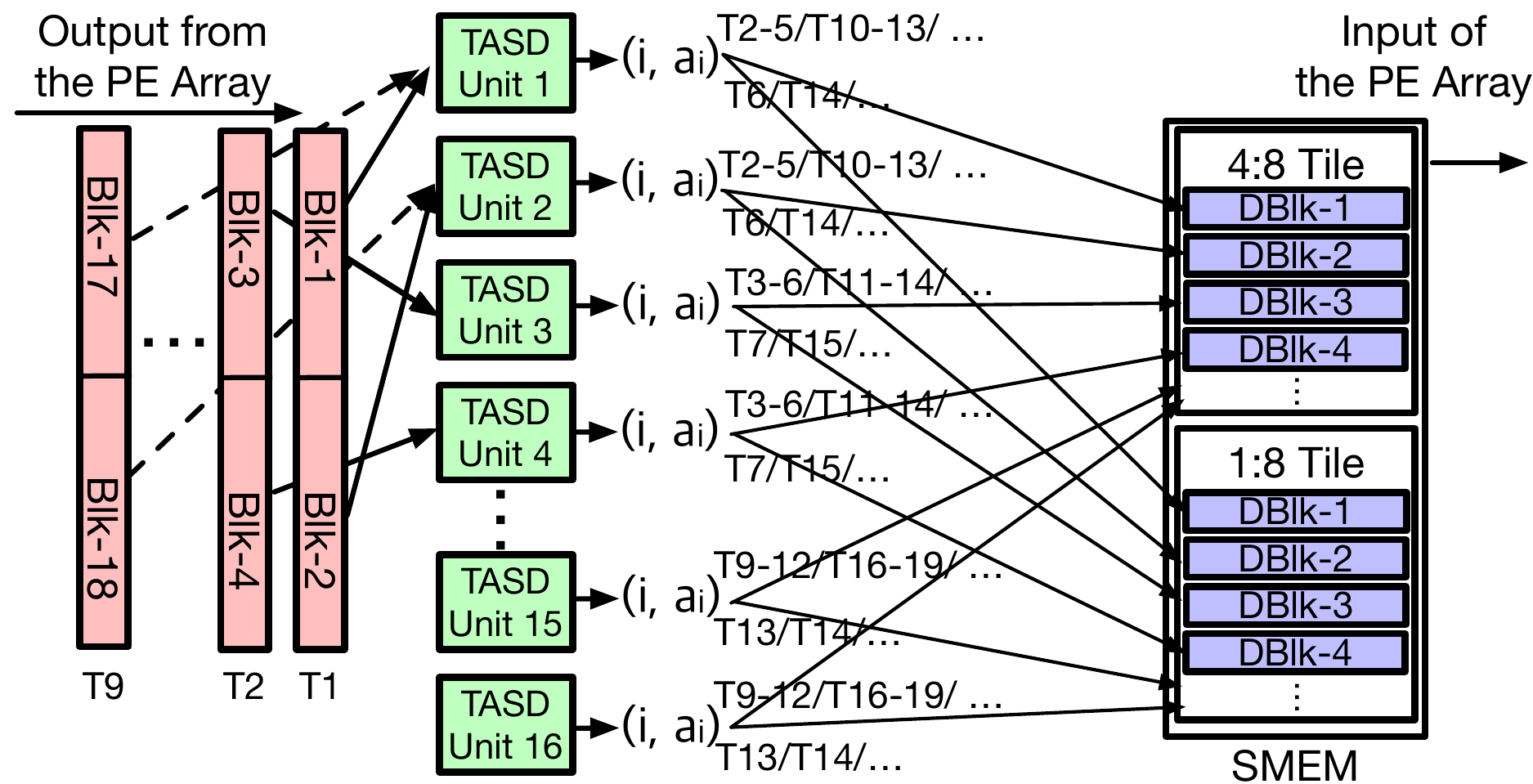}
    \vspace{-1em}
	\caption{Dataflow between PE array and TASD units.}
	\label{fig:tasd-flow}
    \vspace{-2em}
\end{figure}

In \autoref{fig:tasd-hw}, we show the overall design of a TASD HW composed of four TTCs similar to the one used in the previous work \cite{wu2022sparseloop}.
The modification we add on top of the N:M accelerator, such as STC or VEGETA, is the TASD unit (shown in the right part of \autoref{fig:tasd-hw}) that dynamically extracts TASD terms from the activation tensor, similar to the DAP unit in S2TA~\cite{liu2022hpca}.
TASD-W can be applied offline through pre-processing since weights do not change during runtime, but TASD-A requires the TASD unit as activations will be dynamically generated at runtime.

Given the computation latency on TTCs, the minimum number of TASD units per TTC to hide the latency of TASD units depends on the mapping and TTC implementations.
For example, each TTC-VEGETA with M=8 generates 16 (number of PE columns in each TTC) output elements per cycle (i.e. 2 blocks per cycle) as shown in \autoref{fig:tasd-flow}, which will be fed to TASD units. 
For an M-element block, a TASD unit sequentially extracts the largest values, so the decomposition takes up to M cycles 
as the sum of Ns in a TASD configuration cannot be larger than M.

The example in \autoref{fig:tasd-flow} uses TASD configuration composed of 4:8 and 1:8, so it takes 5 cycles per block. 
At T1 (cycle 1), two blocks (Blk-1, Blk-2) will be produced from the PE array, and Blk-1 and Blk-2 will be processed by TASD Unit 1 and TASD Unit 2, respectively (each cycle, two TASD units start execution). 
During T2-T5, Blk-1 and Blk-2 will be used to extract Decomposed Blocks, DBlk-1 and DBlk-2 for 4:8 Tiles.
Then, during T6, DBlk-1 and DBlk-2 for 1:8 Tiles will be generated and stored. The decomposed blocks will be used as the inputs of the next layer.
With 16 TASD units, a TTC-VEGETA can operate without stalls on the PE array due to the decomposition as a TASD unit is always guaranteed to be available after M cycles (i.e. by Little's law, $16 = 2\times8$). 
We present the area overhead for TASD units in \autoref{sec:eval}.


\begin{figure}[!t]
	\centering
	\includegraphics[width=0.45\textwidth]{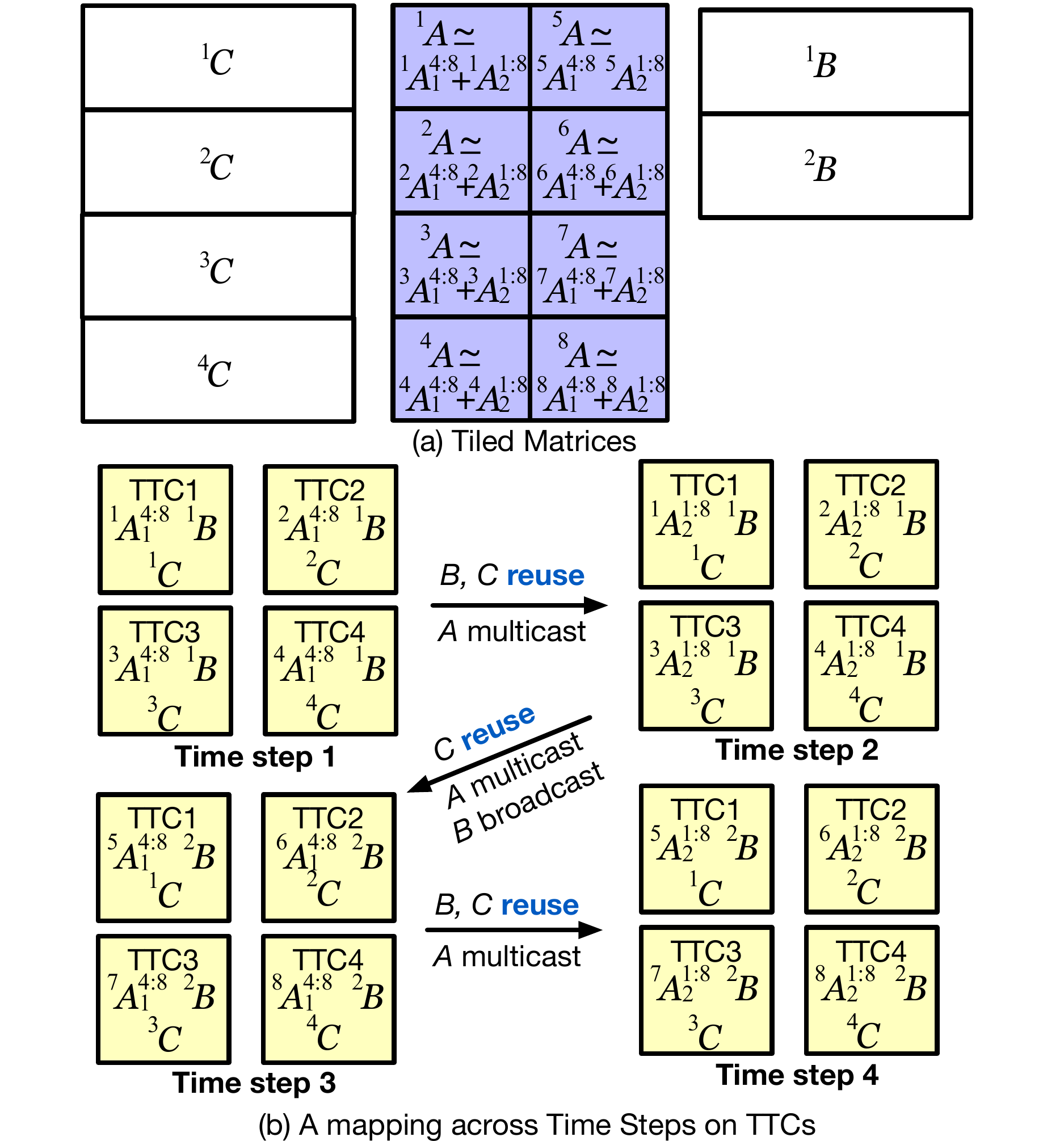}
 \vspace{-1em}
	\caption{A mapping of matrix multiplication on TTCs.
 }
    \vspace{-1em}
	\label{fig:tasd-mapping}
\end{figure}
\begin{figure*}[!t]
    \centering
    \includegraphics[width=0.95\linewidth]{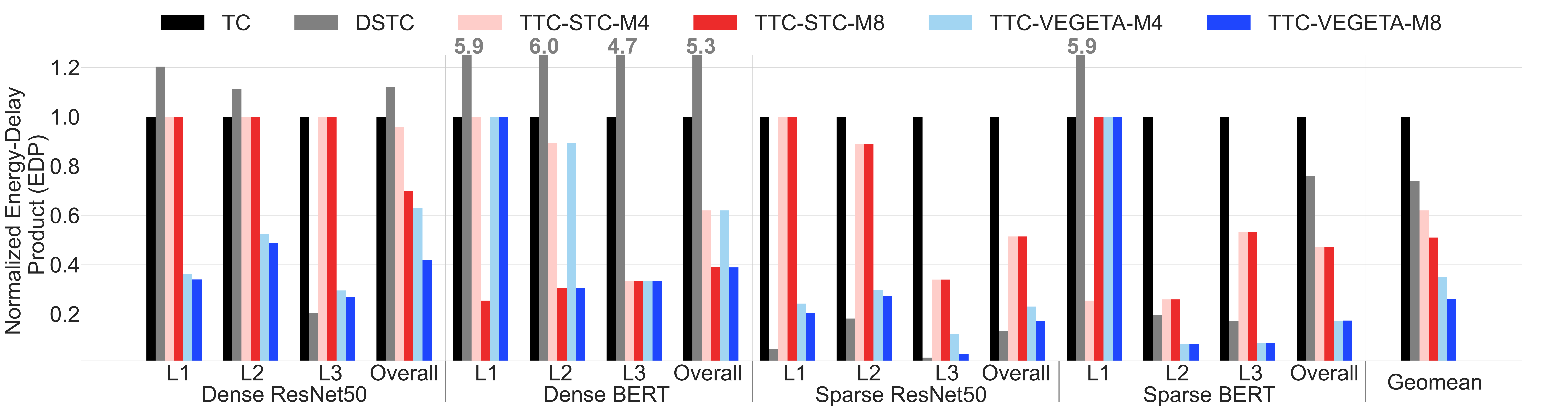}
    \vspace{-0.5em}
    \caption{
    Energy-delay-products for running dense and sparse ResNet50 and BERT. For TTC-STC and TTC-VEGETA, we use the TASD transformations found from TASDER.
    M4/M8 represents design with N:4/N:8 supports.
    }
    \vspace{0.5em}
    \label{fig:dense_sparse_comparison}
\end{figure*}
\textbf{Decomposition-aware dataflow.}
In \autoref{fig:tasd-mapping}, we show a mapping of a matrix multiplication with an approximated matrix A using a TASD configuration, 4:8 and 1:8 for the TTC.
We first show how we tile the matrices and how they are mapped in the private register file and shared buffer of each TTC.
When matrix $A$ is decomposed into two TASD terms, $A_1$, and $A_2$, the original matrix multiplication can be approximated as the sum of the two matrix multiplications and accumulation ($A_1\times B + A_2\times B$). 
As the two matrix operations share the same input $B$ and partial sum $C$, we \textbf{keep B tiles in the L2 Scratchpad Memory (SMEM) and C tiles stationary in the L1 SMEM} of TTC while changing decomposed A tiles to temporally reuse B and C tiles for data reuse (between timestep 1 and 2, timestep 3 and 4 in ~\autoref{fig:tasd-mapping} (b)).
For each accelerator tile, (i.e. for each timestep), we \textbf{keep each element of A tile stationary in the register file of each PE} for the temporal reuse, while the B and C elements are mapped correspondingly. 
By increasing the tile size for GEMM-N dimension, the reuse count for A tile at PE level could increase, which is limited by the size of the capacity of each SMEM.
We swap C tiles at the very end to minimize the number of write-back operations to other levels.
Although we maximize reuses for decomposed tiles, there is still unavoidable overhead such as reading C tiles again, but this is insignificant compared to the energy saving by skipping ineffectual computations using TASD.
In \autoref{subsec:energy_breakdown}, we quantify the energy overhead.






\section{Evaluation}
\label{sec:eval}
\subsection{Methodology}

TASD accelerates both sparse and dense DNNs \textit{without fine-tuning}, so we evaluate TASD-W on sparse DNNs from SparseZoo~\cite{SparseZoo} and TASD-A on dense DNNs from TorchVision~\cite{torchvision}.
We use a classic convolutional network, ResNet50~\cite{he2016resnet}, and a transformer-based network, BERT~\cite{devlin-etal-2019-bert}.  
For the baseline HW, we compare against dense tensor core (TC)~\cite{nvidia_ampere} and dual-side sparse tensor core (DSTC)~\cite{wang:dual-side} as representative dense and unstructured sparse accelerators.
We configure these baselines as in the Sparseloop Artifacts~\cite{sparseloop-artifact}.
We use 4 variants of TTC, based on STC and VEGETA with N:4 and N:8 designs, to show the extra benefits of TASD from the flexibility of the structured sparse hardware as summarized in \autoref{tab:hw}.
All designs use the same memory hierarchy and the same amount of PEs (MACs) to ensure a fair comparison.
We clarify that we do not exploit both sparsities concurrently for skipping computations. We either use TASD-W or TASD-A depending on the workloads since supporting both sparsities requires non-trivial area/energy costs~\cite{wu2021sparseloop, highlight2023micro} and is not compatible to approximate with TASD.


We develop TASDER as a framework to search for TASD transformations and calculate the accuracy of the model with each TASD transformation using PyTorch~\cite{paszke2019pytorch}. 
Following the requirement in MLPerf~\cite{mlperf} inference benchmark, we only consider a model as valid with TASD \textit{if the model achieves an accuracy higher than 99\% of the accuracy of the original model.}
Next, we run each DNN layer with the given TASD series configuration using Sparseloop~\cite{wu2022sparseloop}, a sparse accelerator modeling framework to obtain per-layer results and aggregate the results for the entire network, which is consistent with prior accelerator simulation frameworks~\cite{timeloop, cosa, maestro, zigzag, scale-sim}. Sparseloop estimates the performance and energy consumption of the given HW by analyzing data movements across different memory hierarchy levels and processing elements, and actual computation based on the given dataflow. 
\textbf{We simulate all layers in the networks (marked as ``Overall")}, but to show per-layer results additionally, we choose three representative layers (from early, mid, late) as shown in \autoref{tab:workload}.

\begin{small}
\begin{table}[t]

\begin{center}
\small

\caption{Summary of different HW designs. TASD 1T and 2T indicates using TASD 1 term and 2 terms, respectively.
}
\vspace{1em}
\begin{tabular}{c|c}
{\bf HW Design} & {\bf Sparsity Support from HW} \\
\hline
\hline
TC  & None  \\
\hline
DSTC & Unstructured \\
\hline
TTC-STC-M4 & 2:4 (TASD 1T) \\               
\hline
TTC-STC-M8 & 4:8 (TASD 1T) \\
\hline
TTC-VEGETA-M4 & 
\begin{tabular}[c]{@{}c@{}}1:4, 2:4 (TASD 1T) \\ + 3:4 (TASD 2T)\end{tabular} 
  \\
\hline
TTC-VEGETA-M8 
&\begin{tabular}[c]{@{}c@{}}1:8, 2:8, 4:8 (TASD 1T) \\ + 3:8, 5:8, 6:8 (TASD 2T)\end{tabular}  

\end{tabular}

\label{tab:hw}
\end{center}
\end{table}
\end{small}

\begin{small}
\begin{table}[t]
\begin{center}
\footnotesize

\caption{Representative layers from the target workloads. L1, L2, and L3 are representative layers. RN and Act. refer to ResNet and Activation, respectively.
}
\begin{tabular}{c|c|c|c}
{\bf Model} &{\bf Weight}& {\bf Act.} & {\bf Layers Dimensions}\\
\hline
\hline   

\begin{tabular}[c]{@{}c@{}}Dense RN50 \\ (ReLU-based)\end{tabular}     &Dense& Sparse & 
\begin{tabular}[c]{@{}c@{}}L1: M784-N128-K1152 \\ L2: M3136-N64-K576\end{tabular}
\\
\cline{1-3}
Sparse RN50 &Sparse& Sparse & L3: M196-K2304-N256 \\
    
\hline
\begin{tabular}[c]{@{}c@{}}Dense BERT \\ (GeLU-based)\end{tabular}     &Dense& Dense & 
\begin{tabular}[c]{@{}c@{}}L1: M768-N128-K768 \\ L2: M3072-N128-K768\end{tabular}
\\
\cline{1-3}
Sparse BERT &Sparse& Dense & L3: M768-N128-K3072 \\

\end{tabular}

\label{tab:workload}
\end{center}
\end{table}
\end{small}

We use energy-delay product (EDP), latency, and energy for the metrics. We also provide additional experiment results on the theoretical MACs reductions for other various DNNs, comparison against structured sparse accelerators, and TASD on a real system.




\subsection{DNN acceleration with TASD}

\autoref{fig:dense_sparse_comparison} shows the EDP for the 4 workloads on various DNN accelerators, normalized to the dense TC.
Even though DSTC is able to exploit unstructured sparsity, the overhead of unstructured sparse acceleration (such as accessing accumulation buffer frequently) offsets the benefit and even outweighs the benefits when the workload has only one sparse operand or there is no sparse operand, causing 12\% and 167\% \textbf{larger} EDP for dense ResNet50 and dense BERT while reducing EDP by 55\% for sparse BERT.
DSTC works best for sparse ResNet50 and improves EDP by 87\%, as both weight and activation tensors are unstructured sparse with a high sparsity degree (95\% sparse weight).
Overall, it is able to reduce EDP by 35\% across all workloads on average.


Unlike DSTC, TASD-based TTC accelerators improve EDP over the TC baseline for all workloads. 
With the flexibility in sparsity patterns, TTC-VEGETA-M8 improves EDP for all workloads, by 58\%/61\% for dense ResNet50/BERT and 83\%/82\% for sparse ResNet50/BERT. 
Even with a single fixed pattern, TTC-STC-M4 improves by 4\%/32\% for dense ResNet50/BERT and 49\%/53\% for sparse ResNet50/BERT. 
This result shows that TASD can effectively leverage structured sparse hardware for off-the-shelf dense and sparse DNNs with no fine-tuning, and the extra flexibility (increasing M) in the baseline accelerator increases the benefit.

\begin{figure}[t]
	\centering
	\includegraphics[width=0.49\textwidth]{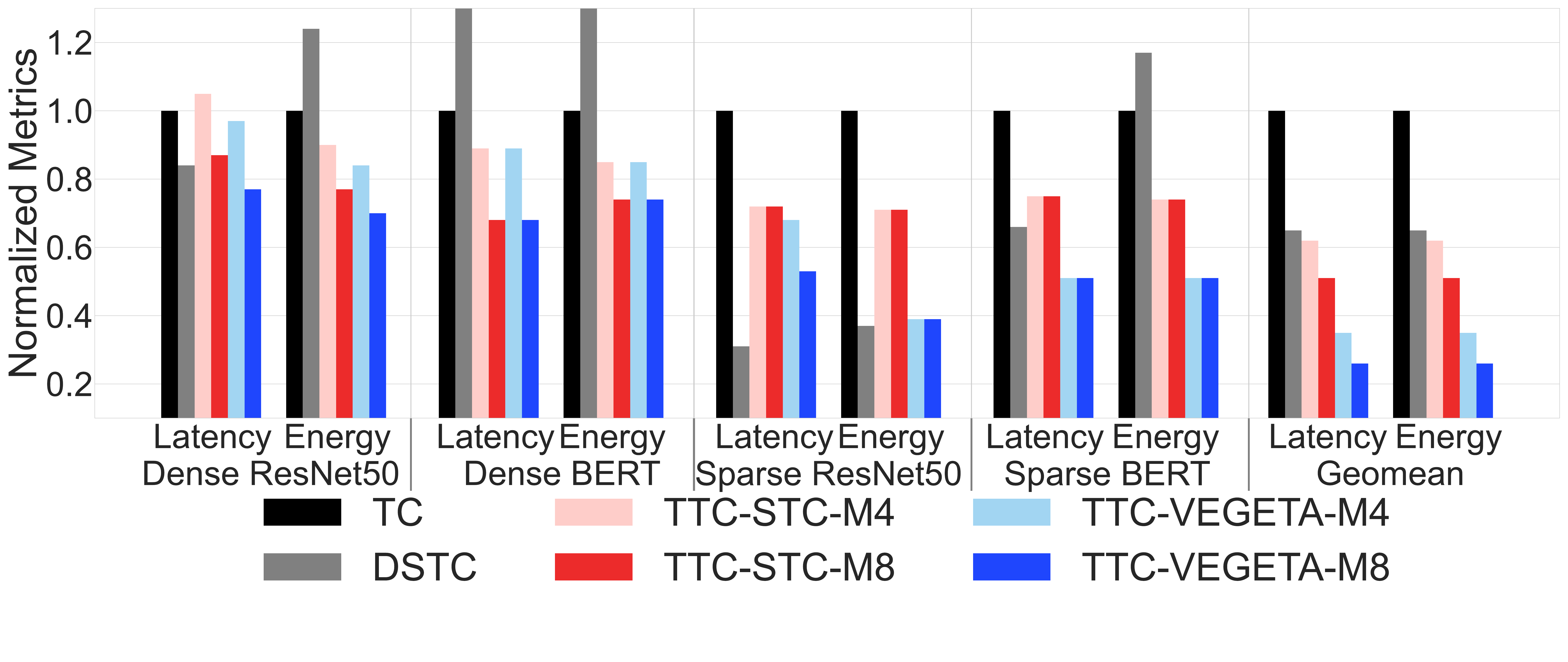}
        \vspace{-3em}
	\caption{
        Latency and energy for various designs.
 }
	\label{fig:overall_energy_latency_edp}
 \end{figure}

\autoref{fig:overall_energy_latency_edp} provides end-to-end latency and energy consumption for various designs. TTC-VEGETA-M8 is the most energy-efficient across all workloads and is slightly slower than DSTC only for sparse ResNet50. 
This result shows TASD provides a better overall tradeoff than unstructured sparse accelerators, considering their high area overheads.

  \begin{figure}[t]
	\centering
	\includegraphics[width=0.49\textwidth]{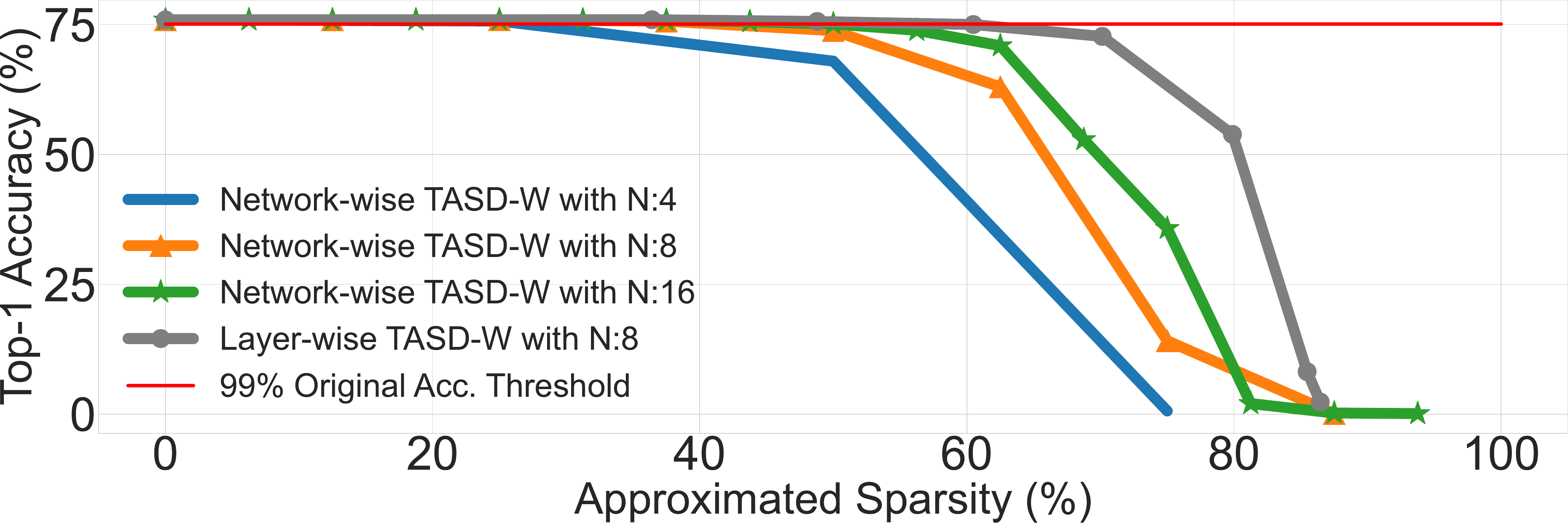}
 	\includegraphics[width=0.49\textwidth]{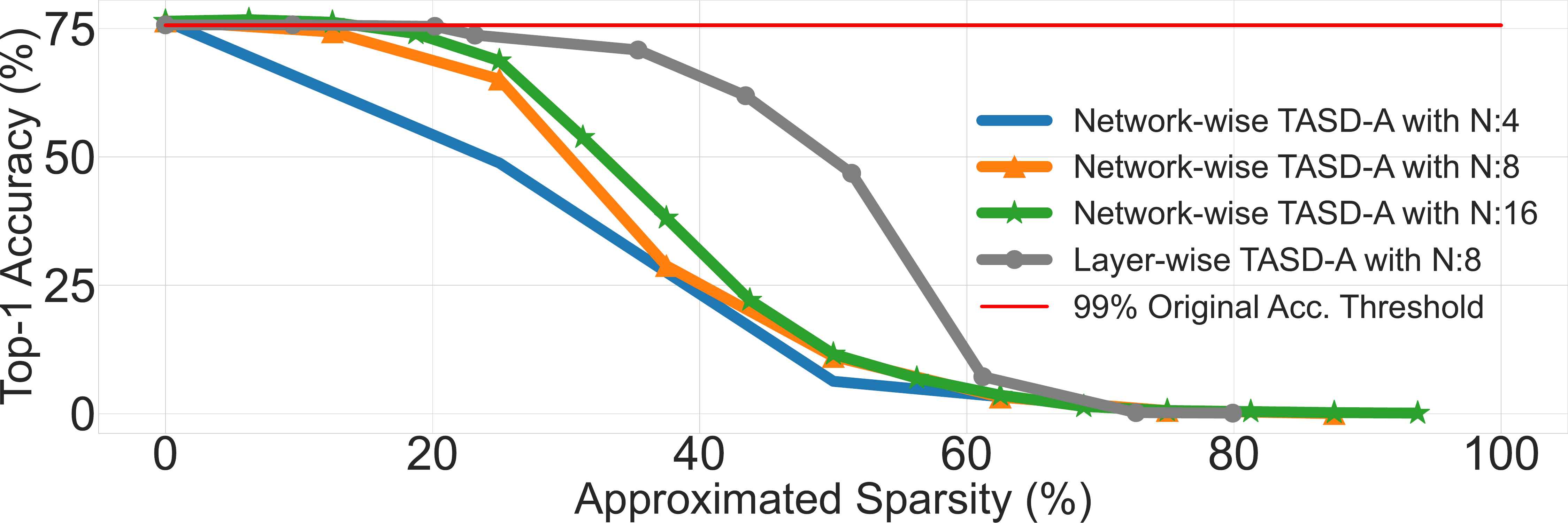}
        \vspace{-1em}
	\caption{
Network-wise and Layer-wise TASD on ResNet50. Upper: TASD-W. Lower: TASD-A.
 }
        \vspace{-1em}
	\label{fig:resnet50-95-networkwise-w}
\end{figure}
\subsection{Analysis of TASD} 
\paragraph{Network-wise vs. layer-wise TASD.}
The upper plot of \autoref{fig:resnet50-95-networkwise-w} shows the impact of network-wise TASD-W on the top-1 accuracy of unstructured sparse ResNet50 (95\% sparsity).
We applied network-wise TASD-W with N:4, N:8, and N:16 structured sparsities. 
For example, the network-wise TASD-W with 2:4 uses one TASD term with the 2:4 pattern to the weights of all convolution and fully-connected layers in the sparse ResNet50. 
Since TASD is a lossy method, aggressive TASD series approximation can result in a notable accuracy drop.
Among different N:4, N:8, N:16 options, we found that 3:4, 5:8, 10:16 is the most aggressive approximation among the available options while meeting the 99\% accuracy requirement. 
Especially, using network-wise TASD-W 5:8 (4:8 + 1:8 for TTC-VEGETA) and gating the compute units for sparse activations, compared to the dense baseline, we observe it achieves 24\% and 53\% reduction in cycle and energy respectively, thus reducing 75\% EDP for Sparse ResNet50. 




Using different TASD series configurations for different layers is more effective as it can adjust the aggressiveness for each layer. 
To choose a TASD configuration per layer, we use the sparsity-based selection method that we introduce in \autoref{sec:codesgin}. 
By changing the hyperparameter, alpha, we are able to adjust the aggressiveness of our approximation method.
As layer-wise TASD-W can be adaptive to each layer, the overall approximation can be applied more aggressively.
As a result, compared to the dense baseline, we observe 47\% and 61\% reduction in energy and cycle, respectively, thus reducing 83\% EDP for Sparse ResNet50 as shown in \autoref{fig:dense_sparse_comparison}.

In the lower plot of \autoref{fig:resnet50-95-networkwise-w}, we show the top-1 accuracy when network-wise and layer-wise TASD-A is applied with different TASD series.
Similar to TASD-W, layer-wise TASD-A is more effective than network-wise TASD-A.
However, the accuracy loss due to approximation shows up with a much smaller approximated sparsity.
As shown earlier in ~\autoref{fig:resnet50_layer_sparsity}, the sparsity degree is larger in weights compared to that in activations for sparse ResNet50.
Thus, the same TASD series drops a larger portion of non-zeros in TASD-A than TASD-W, incurring a higher loss of accuracy.



\subsection{
Energy and area overhead due to TASD
}
\label{subsec:energy_breakdown}
\begin{figure}[!t]
	\centering
	\includegraphics[width=0.48\textwidth,]{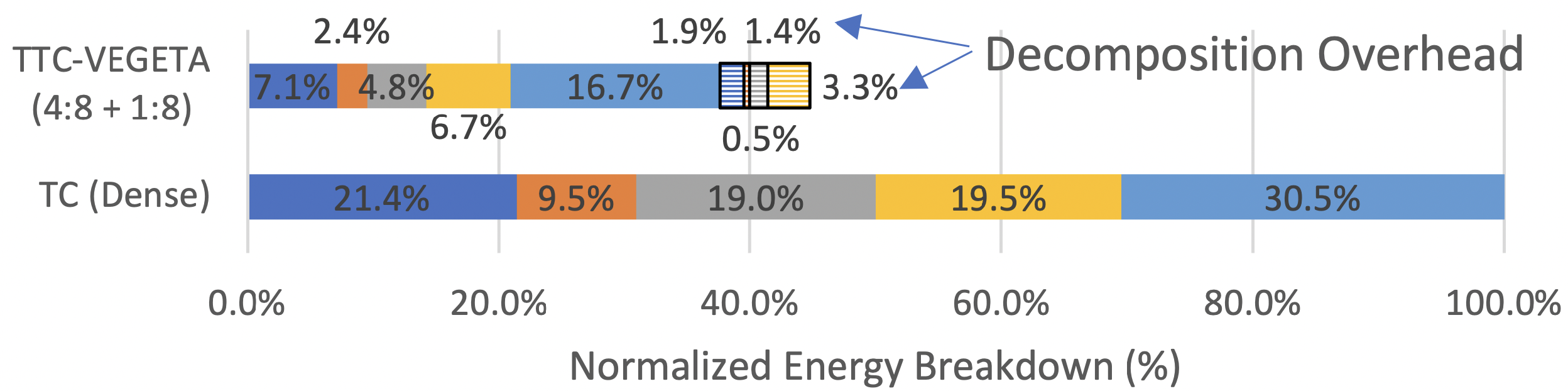}
        \vspace{-1.5em}
	\caption{
        Energy Breakdown: TTC vs. Dense TC.}
        \vspace{-0.5em}
        
	\label{fig:energy_breakdown}
 \end{figure}
\begin{figure}[!t]
	\centering
	\includegraphics[width=0.48\textwidth]{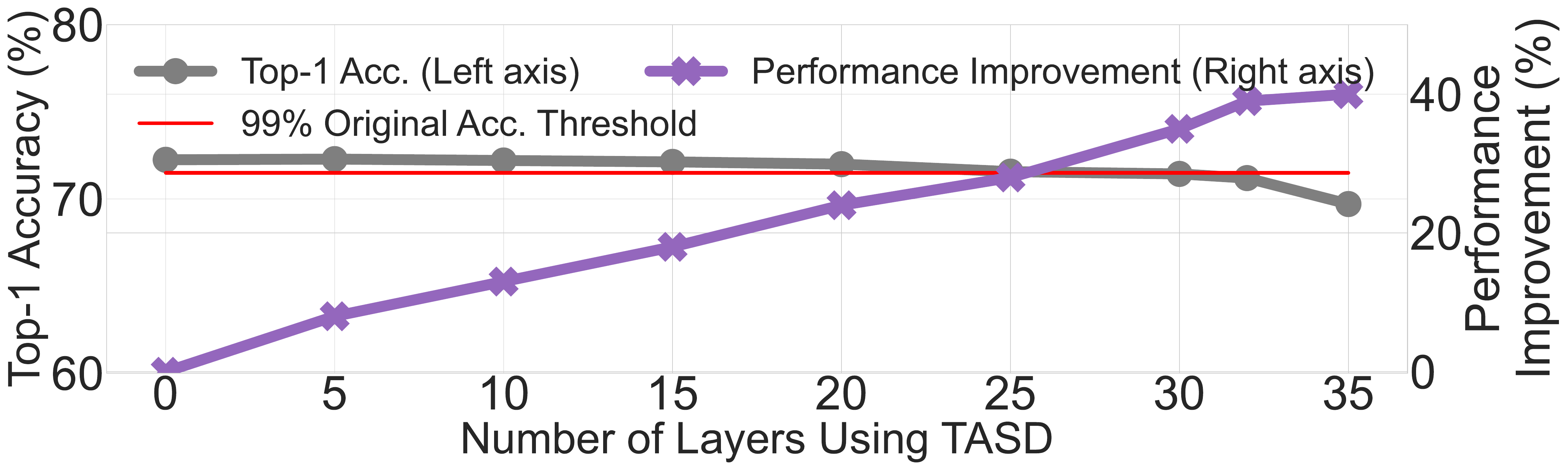}
	        \vspace{-0.5em}
    \caption{
        TASD-W on NVIDIA RTX3080 GPU with Sparse ResNet34.}
        \vspace{-0.5em}
	\label{fig:gpueval}
 \end{figure}
 
\autoref{fig:energy_breakdown} shows the energy breakdown for a representative layer from sparse ResNet50 for dense TC and TTC-VEGETA with a TASD configuration of 4:8+1:8. TTC-VEGETA exploits sparsity and saves energy at all levels of the architecture, which saves 55\% energy over the dense TC. Moreover, the decomposition-aware dataflow in \autoref{sec:codesgin} minimizes decomposition overheads by accessing the RF (with C reuse) and SMEM (with B reuse) instead of accessing DRAM.

We measured the area overhead to support TASD on top of the existing structured sparse HW (the TASD units) through RTL prototyping and synthesis with Nangate 15nm Technology Library. We observe up to 2\% of the area for all PEs as TASD units are composed of simple comparator trees.

\subsection{Using TASD on a real system}
To validate the effectiveness of TASD, we evaluate TASD on a real system.
First, we use our framework, TASDER, to find the TASD-W configuration that meets the accuracy constraint with an unstructured sparse ResNet34 model from SparseZoo \cite{SparseZoo}. For the accuracy evaluation, we use ImageNet as the dataset. Next, we export the model using ONNX format. 
To properly measure the inference latency of DNNs exported with ONNX format, we use TensorRT, the state-of-the-art deep learning inference runtime maintained by NVIDIA. We build TensorRT engines for both the baseline model and the model with TASD-W exported in the previous step. Finally, we execute the engine with (3, 224, 224) inputs, with a batch size of 32-128, and measure the latency for each model. We achieved 28\%/39\% speed-up compared to the dense model execution with 0.9\%/1.5\% accuracy drop as shown in \autoref{fig:gpueval}. We also evaluated TASD-W with ResNet50, ResNet101, BERT, and observed up to 25\% speed-up.

\section{Related Work}
\subsection{SW techniques for structured sparse HW}

\textbf{Solutions with fine-tuning.}
DominoSearch \cite{sun2021domino} proposed a method to find layer-wise N:M sparsity during training. 
Optimal N:M~\cite{chmiel2022optimal} enforces the structured pattern during training and applies the structured pattern to input activations.
This line of work is orthogonal to our work as we focus on approximating unstructured sparsity without fine-tuning.
Fine-tuning will increase the benefit of TASD, as more aggressive approximation can now maintain the same model accuracy. 
Doping~\cite{thakker2021doping} uses an extremely sparse matrix in addition to a compressed matrix derived from Kronecker products to improve the quality of the model, but unlike TASD, it uses the extra extremely sparse matrix to give additional freedom during the training.

\textbf{Solutions without fine-tuning.}
SparseTIR~\cite{ye2023sparsetir} introduces composable formats and transformations for sparse compilation of DL workloads.
However, they have not considered approximating sparse tensors and accelerating DNNs using structured sparse hardware. Another work~\cite{pool2021channel} shows permuting channels in the weight tensors can recover accuracy easily when training N:M sparse networks. 
TASD is compatible with channel permutation, and we believe combining these two orthogonal techniques will further improve the accuracy of decomposed models with approximation.

\subsection{HW support for sparse DNNs}

\textbf{Sparsity support for DNN inference.}
Different architectures have been proposed to support for weight sparsity~\cite{eyerissv2, nvidia_ampere, rasa, eureka2023micro},
for activation sparsity~\cite{jang2021samsung},
and more recently, for both~\cite{wang:dual-side, rmstc2023micro, highlight2023micro}.
As mentioned earlier, unstructured sparse HW provides native support for any sparsity pattern but is more costly to build; structured sparse HW is efficient but requires model fine-tuning.
TASD bridges the gap by providing an unstructured sparse interface while only requiring structured sparse HW.

\textbf{Sparsity support for DNN training.}
Since weight tensors are mostly dense during training, prior work has focused on activation and gradient sparsity during DNN training.
A simpler support is to compress sparse activation and gradient, such as CompressDMA~\cite{rhu2018cdma} and ZComp~\cite{akin2019zcomp}. These techniques save data movements and memory requirements, but not overall compute.
The more complex techniques target reducing computation during training, such as TensorDash~\cite{mahmoud2020tensordash} and SAVE~\cite{gong2020save}. However, they need to give up data movement savings for better support for sparse tensor transposition during training.
TASD can potentially be used to approximate sparse activations and gradients, but we leave this to future work. 

%


\section{Conclusion}
Sparse DNN model developers prefer to induce unstructured sparsity for expressibility, while sparse DNN hardware designers prefer to support structured sparsity for HW efficiency.
This mismatch of the desired sparsity pattern 
between different parties
prevents sparse DNN acceleration from being widely adopted in practice.
To close the gap,
we introduce TASD, a method that approximates an unstructured sparse tensor with a series of structured sparse tensors.
Next, we propose a framework, TASDER, which finds TASD configuration for each DNN layer
to accelerate sparse and dense DNNs.
To maximize the benefit of TASD,
we propose a simple architectural extension and dataflow on top of structured sparse accelerators.
TASD improves EDP up to 83\% and 74\% on average, while maintaining 99\% of the model accuracy without any fine-tuning.
We also show it achieves up to 39\% performance improvement on a real system evaluation.


\section*{Acknowledgements}

We thank Jeff Pool, Angshuman Parashar, Michael Pellauer, Qijing Huang, Joel Emer, Christopher Hughes, Hyesoon Kim, and Vivek Sarkar for their valuable feedback and insightful discussions. 

\newpage
\bibliography{example_paper}

\begin{thebibliography}{73}
\providecommand{\natexlab}[1]{#1}
\providecommand{\url}[1]{\texttt{#1}}
\expandafter\ifx\csname urlstyle\endcsname\relax
  \providecommand{\doi}[1]{doi: #1}\else
  \providecommand{\doi}{doi: \begingroup \urlstyle{rm}\Url}\fi

\bibitem[Akin et~al.(2019)Akin, Chishti, and Alameldeen]{akin2019zcomp}
Akin, B., Chishti, Z.~A., and Alameldeen, A.~R.
\newblock Zcomp: Reducing dnn cross-layer memory footprint using vector extensions.
\newblock In \emph{Proceedings of the 52nd Annual IEEE/ACM International Symposium on Microarchitecture}, MICRO '52, pp.\  126–138, New York, NY, USA, 2019. Association for Computing Machinery.
\newblock ISBN 9781450369381.
\newblock \doi{10.1145/3352460.3358305}.
\newblock URL \url{https://doi.org/10.1145/3352460.3358305}.

\bibitem[AMD(2023)]{AMD_MI300}
AMD.
\newblock "amd instinct mi300 instruction set architecture reference guide", 2023.
\newblock \url{https://www.amd.com/content/dam/amd/en/documents/instinct-tech-docs/instruction-set-architectures/amd-instinct-mi300-cdna3-instruction-set-architecture.pdf}.

\bibitem[Bambhaniya et~al.(2023)Bambhaniya, Yazdanbakhsh, Subramanian, and Krishna]{bambhaniya2023accelerating}
Bambhaniya, A.~R., Yazdanbakhsh, A., Subramanian, S., and Krishna, T.
\newblock Accelerating attention based models via {HW}-{SW} co-design using fine-grained sparsification.
\newblock In \emph{Architecture and System Support for Transformer Models (ASSYST @ISCA 2023)}, 2023.

\bibitem[Bambhaniya et~al.(2024)Bambhaniya, Yazdanbakhsh, Subramanian, Kao, Agrawal, Evci, and Krishna]{bambhaniya2024progressive}
Bambhaniya, A.~R., Yazdanbakhsh, A., Subramanian, S., Kao, S.-C., Agrawal, S., Evci, U., and Krishna, T.
\newblock Progressive gradient flow for robust n:m sparsity training in transformers, 2024.

\bibitem[Chen et~al.(2019)Chen, Yang, Emer, and Sze]{eyerissv2}
Chen, Y.-H., Yang, T.-J., Emer, J., and Sze, V.
\newblock Eyeriss v2: A flexible accelerator for emerging deep neural networks on mobile devices.
\newblock \emph{IEEE Journal on Emerging and Selected Topics in Circuits and Systems}, 9\penalty0 (2):\penalty0 292--308, 2019.

\bibitem[Chmiel et~al.(2022)Chmiel, Hubara, Banner, and Soudry]{chmiel2022optimal}
Chmiel, B., Hubara, I., Banner, R., and Soudry, D.
\newblock Optimal fine-grained n:m sparsity for activations and neural gradients.
\newblock \emph{arXiv preprint arXiv:2203.10991}, 2022.

\bibitem[Chowdhery et~al.(2022)Chowdhery, Narang, Devlin, Bosma, Mishra, Roberts, Barham, Chung, Sutton, Gehrmann, et~al.]{chowdhery2022palm}
Chowdhery, A., Narang, S., Devlin, J., Bosma, M., Mishra, G., Roberts, A., Barham, P., Chung, H.~W., Sutton, C., Gehrmann, S., et~al.
\newblock Palm: Scaling language modeling with pathways.
\newblock \emph{arXiv preprint arXiv:2204.02311}, 2022.

\bibitem[Dao et~al.(2021)Dao, Chen, Liang, Yang, Song, Rudra, and Re]{dao2021pixelated}
Dao, T., Chen, B., Liang, K., Yang, J., Song, Z., Rudra, A., and Re, C.
\newblock Pixelated butterfly: Simple and efficient sparse training for neural network models.
\newblock \emph{arXiv preprint arXiv:2112.00029}, 2021.

\bibitem[Deng et~al.(2009)Deng, Dong, Socher, Li, Li, and Fei-Fei]{deng2009imagenet}
Deng, J., Dong, W., Socher, R., Li, L.-J., Li, K., and Fei-Fei, L.
\newblock Imagenet: A large-scale hierarchical image database.
\newblock In \emph{2009 IEEE conference on computer vision and pattern recognition}, pp.\  248--255. Ieee, 2009.

\bibitem[Devlin et~al.(2019)Devlin, Chang, Lee, and Toutanova]{devlin-etal-2019-bert}
Devlin, J., Chang, M.-W., Lee, K., and Toutanova, K.
\newblock {BERT}: Pre-training of deep bidirectional transformers for language understanding.
\newblock In \emph{Proceedings of the 2019 Conference of the North {A}merican Chapter of the Association for Computational Linguistics: Human Language Technologies, Volume 1 (Long and Short Papers)}, pp.\  4171--4186, Minneapolis, Minnesota, June 2019. Association for Computational Linguistics.
\newblock \doi{10.18653/v1/N19-1423}.

\bibitem[Dosovitskiy et~al.(2020)Dosovitskiy, Beyer, Kolesnikov, Weissenborn, Zhai, Unterthiner, Dehghani, Minderer, Heigold, Gelly, et~al.]{vit}
Dosovitskiy, A., Beyer, L., Kolesnikov, A., Weissenborn, D., Zhai, X., Unterthiner, T., Dehghani, M., Minderer, M., Heigold, G., Gelly, S., et~al.
\newblock An image is worth 16x16 words: Transformers for image recognition at scale.
\newblock \emph{arXiv preprint arXiv:2010.11929}, 2020.

\bibitem[Fang et~al.(2022)Fang, Zhou, and Wang]{fang2022algorithm}
Fang, C., Zhou, A., and Wang, Z.
\newblock An algorithm--hardware co-optimized framework for accelerating n: M sparse transformers.
\newblock \emph{IEEE Transactions on Very Large Scale Integration (VLSI) Systems}, 30\penalty0 (11):\penalty0 1573--1586, 2022.

\bibitem[Fedus et~al.(2021)Fedus, Zoph, and Shazeer]{fedus2021switch}
Fedus, W., Zoph, B., and Shazeer, N.
\newblock Switch transformers: Scaling to trillion parameter models with simple and efficient sparsity.
\newblock \emph{arXiv preprint arXiv:2101.03961}, 2021.

\bibitem[Frantar \& Alistarh(2023)Frantar and Alistarh]{sparsegpt}
Frantar, E. and Alistarh, D.
\newblock Massive language models can be accurately pruned in one-shot.
\newblock \emph{arXiv preprint arXiv:2301.00774}, 2023.

\bibitem[Gondimalla et~al.(2023)Gondimalla, Thottethodi, and Vijaykumar]{eureka2023micro}
Gondimalla, A., Thottethodi, M., and Vijaykumar, T.~N.
\newblock Eureka: Efficient tensor cores for one-sided unstructured sparsity in dnn inference.
\newblock In \emph{Proceedings of the 56th Annual IEEE/ACM International Symposium on Microarchitecture}, MICRO '23, pp.\  324–337, New York, NY, USA, 2023. Association for Computing Machinery.
\newblock ISBN 9798400703294.
\newblock \doi{10.1145/3613424.3614312}.

\bibitem[Gong et~al.(2020)Gong, Ji, Fletcher, Hughes, Baghsorkhi, and Torrellas]{gong2020save}
Gong, Z., Ji, H., Fletcher, C.~W., Hughes, C.~J., Baghsorkhi, S., and Torrellas, J.
\newblock Save: Sparsity-aware vector engine for accelerating dnn training and inference on cpus.
\newblock In \emph{2020 53rd Annual IEEE/ACM International Symposium on Microarchitecture (MICRO)}, pp.\  796--810, 2020.
\newblock \doi{10.1109/MICRO50266.2020.00070}.

\bibitem[Han et~al.(2015)Han, Mao, and Dally]{han2015deep}
Han, S., Mao, H., and Dally, W.~J.
\newblock Deep compression: Compressing deep neural networks with pruning, trained quantization and huffman coding.
\newblock \emph{arXiv preprint arXiv:1510.00149}, 2015.

\bibitem[He et~al.(2016)He, Zhang, Ren, and Sun]{he2016resnet}
He, K., Zhang, X., Ren, S., and Sun, J.
\newblock Deep residual learning for image recognition.
\newblock In \emph{Proceedings of the IEEE conference on computer vision and pattern recognition}, pp.\  770--778, 2016.

\bibitem[Hegde et~al.(2019)Hegde, Asghari-Moghaddam, Pellauer, Crago, Jaleel, Solomonik, Emer, and Fletcher]{extensor}
Hegde, K., Asghari-Moghaddam, H., Pellauer, M., Crago, N., Jaleel, A., Solomonik, E., Emer, J., and Fletcher, C.~W.
\newblock Extensor: An accelerator for sparse tensor algebra.
\newblock In \emph{Proceedings of the 52nd Annual IEEE/ACM International Symposium on Microarchitecture}, MICRO '52, pp.\  319–333, New York, NY, USA, 2019. Association for Computing Machinery.
\newblock ISBN 9781450369381.
\newblock \doi{10.1145/3352460.3358275}.

\bibitem[Hendrycks \& Gimpel(2016)Hendrycks and Gimpel]{gelu}
Hendrycks, D. and Gimpel, K.
\newblock Gaussian error linear units (gelus).
\newblock \emph{arXiv preprint arXiv:1606.08415}, 2016.

\bibitem[Howard et~al.(2017)Howard, Zhu, Chen, Kalenichenko, Wang, Weyand, Andreetto, and Adam]{mobilenets}
Howard, A.~G., Zhu, M., Chen, B., Kalenichenko, D., Wang, W., Weyand, T., Andreetto, M., and Adam, H.
\newblock Mobilenets: Efficient convolutional neural networks for mobile vision applications.
\newblock \emph{arXiv preprint arXiv:1704.04861}, 2017.

\bibitem[Huang et~al.(2023)Huang, Wang, Tsai, Zhang, Ding, and Xie]{rmstc2023micro}
Huang, G., Wang, Z., Tsai, P.-A., Zhang, C., Ding, Y., and Xie, Y.
\newblock Rm-stc: Row-merge dataflow inspired gpu sparse tensor core for energy-efficient sparse acceleration.
\newblock In \emph{Proceedings of the 56th Annual IEEE/ACM International Symposium on Microarchitecture}, MICRO '23, pp.\  338–352, New York, NY, USA, 2023. Association for Computing Machinery.
\newblock ISBN 9798400703294.
\newblock \doi{10.1145/3613424.3623775}.

\bibitem[Huang et~al.(2021)Huang, Kang, Dinh, Norell, Kalaiah, Demmel, Wawrzynek, and Shao]{cosa}
Huang, Q., Kang, M., Dinh, G., Norell, T., Kalaiah, A., Demmel, J., Wawrzynek, J., and Shao, Y.~S.
\newblock Cosa: Scheduling by constrained optimization for spatial accelerators.
\newblock In \emph{2021 ACM/IEEE 48th Annual International Symposium on Computer Architecture (ISCA)}, pp.\  554--566, 2021.
\newblock \doi{10.1109/ISCA52012.2021.00050}.

\bibitem[Hubara et~al.(2021)Hubara, Chmiel, Island, Banner, Naor, and Soudry]{hubara2021accelerated}
Hubara, I., Chmiel, B., Island, M., Banner, R., Naor, J., and Soudry, D.
\newblock Accelerated sparse neural training: A provable and efficient method to find n: m transposable masks.
\newblock \emph{Advances in neural information processing systems}, 34:\penalty0 21099--21111, 2021.

\bibitem[Jang et~al.(2021)Jang, Lee, Kim, Park, Ardestani, Choi, Kim, Kim, Yu, Abdel-Aziz, Park, Lee, Lee, Kim, Jung, Nam, Lim, Lee, Song, Kwon, Hassoun, Lim, and Choi]{jang2021samsung}
Jang, J.-W., Lee, S., Kim, D., Park, H., Ardestani, A.~S., Choi, Y., Kim, C., Kim, Y., Yu, H., Abdel-Aziz, H., Park, J.-S., Lee, H., Lee, D., Kim, M.~W., Jung, H., Nam, H., Lim, D., Lee, S., Song, J.-H., Kwon, S., Hassoun, J., Lim, S., and Choi, C.
\newblock Sparsity-aware and re-configurable npu architecture for samsung flagship mobile soc.
\newblock In \emph{2021 ACM/IEEE 48th Annual International Symposium on Computer Architecture (ISCA)}, pp.\  15--28, 2021.
\newblock \doi{10.1109/ISCA52012.2021.00011}.

\bibitem[Jeong et~al.(2021)Jeong, Qin, Samajdar, Hughes, Subramoney, Kim, and Krishna]{rasa}
Jeong, G., Qin, E., Samajdar, A., Hughes, C.~J., Subramoney, S., Kim, H., and Krishna, T.
\newblock Rasa: Efficient register-aware systolic array matrix engine for cpu.
\newblock In \emph{2021 58th ACM/IEEE Design Automation Conference (DAC)}, pp.\  253--258, 2021.
\newblock \doi{10.1109/DAC18074.2021.9586257}.

\bibitem[Jeong et~al.(2023)Jeong, Damani, Bambhaniya, Qin, Hughes, Subramoney, Kim, and Krishna]{jeong:vegeta}
Jeong, G., Damani, S., Bambhaniya, A.~R., Qin, E., Hughes, C.~J., Subramoney, S., Kim, H., and Krishna, T.
\newblock Vegeta: Vertically-integrated extensions for sparse/dense gemm tile acceleration on cpus.
\newblock In \emph{2023 IEEE International Symposium on High Performance Computer Architecture (HPCA)}, 2023.

\bibitem[Jouppi et~al.(2017)Jouppi, Young, Patil, Patterson, Agrawal, Bajwa, Bates, Bhatia, Boden, Borchers, Boyle, Cantin, Chao, Clark, Coriell, Daley, Dau, Dean, Gelb, Ghaemmaghami, Gottipati, Gulland, Hagmann, Ho, Hogberg, Hu, Hundt, Hurt, Ibarz, Jaffey, Jaworski, Kaplan, Khaitan, Killebrew, Koch, Kumar, Lacy, Laudon, Law, Le, Leary, Liu, Lucke, Lundin, MacKean, Maggiore, Mahony, Miller, Nagarajan, Narayanaswami, Ni, Nix, Norrie, Omernick, Penukonda, Phelps, Ross, Ross, Salek, Samadiani, Severn, Sizikov, Snelham, Souter, Steinberg, Swing, Tan, Thorson, Tian, Toma, Tuttle, Vasudevan, Walter, Wang, Wilcox, and Yoon]{jouppi2017tpu}
Jouppi, N.~P., Young, C., Patil, N., Patterson, D., Agrawal, G., Bajwa, R., Bates, S., Bhatia, S., Boden, N., Borchers, A., Boyle, R., Cantin, P.-l., Chao, C., Clark, C., Coriell, J., Daley, M., Dau, M., Dean, J., Gelb, B., Ghaemmaghami, T.~V., Gottipati, R., Gulland, W., Hagmann, R., Ho, C.~R., Hogberg, D., Hu, J., Hundt, R., Hurt, D., Ibarz, J., Jaffey, A., Jaworski, A., Kaplan, A., Khaitan, H., Killebrew, D., Koch, A., Kumar, N., Lacy, S., Laudon, J., Law, J., Le, D., Leary, C., Liu, Z., Lucke, K., Lundin, A., MacKean, G., Maggiore, A., Mahony, M., Miller, K., Nagarajan, R., Narayanaswami, R., Ni, R., Nix, K., Norrie, T., Omernick, M., Penukonda, N., Phelps, A., Ross, J., Ross, M., Salek, A., Samadiani, E., Severn, C., Sizikov, G., Snelham, M., Souter, J., Steinberg, D., Swing, A., Tan, M., Thorson, G., Tian, B., Toma, H., Tuttle, E., Vasudevan, V., Walter, R., Wang, W., Wilcox, E., and Yoon, D.~H.
\newblock In-datacenter performance analysis of a tensor processing unit.
\newblock \emph{SIGARCH Comput. Archit. News}, 45\penalty0 (2):\penalty0 1–12, jun 2017.
\newblock ISSN 0163-5964.
\newblock \doi{10.1145/3140659.3080246}.

\bibitem[Krizhevsky et~al.(2012)Krizhevsky, Sutskever, and Hinton]{krizhevsky2012alexnet}
Krizhevsky, A., Sutskever, I., and Hinton, G.~E.
\newblock Imagenet classification with deep convolutional neural networks.
\newblock \emph{Advances in neural information processing systems}, 25, 2012.

\bibitem[Kwon et~al.(2020)Kwon, Chatarasi, Sarkar, Krishna, Pellauer, and Parashar]{maestro}
Kwon, H., Chatarasi, P., Sarkar, V., Krishna, T., Pellauer, M., and Parashar, A.
\newblock Maestro: A data-centric approach to understand reuse, performance, and hardware cost of dnn mappings.
\newblock \emph{IEEE Micro}, 40\penalty0 (3):\penalty0 20--29, 2020.
\newblock \doi{10.1109/MM.2020.2985963}.

\bibitem[Li et~al.(2023)Li, You, Bhojanapalli, Li, Rawat, Reddi, Ye, chyan Chern, Yu, Guo, and Kumar]{li:transformer-sparsity}
Li, Z., You, C., Bhojanapalli, S., Li, D., Rawat, A.~S., Reddi, S., Ye, K., chyan Chern, F.~R., Yu, F., Guo, R., and Kumar, S.
\newblock On emergence of activation sparsity in trained transformers.
\newblock In \emph{2023 International Conference on Learning Representations (ICLR)}, 2023.

\bibitem[Liu et~al.(2022{\natexlab{a}})Liu, Mao, Wu, Feichtenhofer, Darrell, and Xie]{convnext}
Liu, Z., Mao, H., Wu, C.-Y., Feichtenhofer, C., Darrell, T., and Xie, S.
\newblock A convnet for the 2020s.
\newblock In \emph{Proceedings of the IEEE/CVF Conference on Computer Vision and Pattern Recognition}, pp.\  11976--11986, 2022{\natexlab{a}}.

\bibitem[Liu et~al.(2020)Liu, Whatmough, and Mattina]{liu2020sta}
Liu, Z.-G., Whatmough, P.~N., and Mattina, M.
\newblock Systolic tensor array: An efficient structured-sparse gemm accelerator for mobile cnn inference.
\newblock \emph{IEEE Computer Architecture Letters}, 19\penalty0 (1):\penalty0 34--37, 2020.
\newblock \doi{10.1109/LCA.2020.2979965}.

\bibitem[Liu et~al.(2021)Liu, Whatmough, Zhu, and Mattina]{liu2021s2ta}
Liu, Z.-G., Whatmough, P.~N., Zhu, Y., and Mattina, M.
\newblock S2ta: Exploiting structured sparsity for energy-efficient mobile cnn acceleration.
\newblock \emph{arXiv preprint arXiv:2107.07983}, 2021.

\bibitem[Liu et~al.(2022{\natexlab{b}})Liu, Whatmough, Zhu, and Mattina]{liu2022hpca}
Liu, Z.-G., Whatmough, P.~N., Zhu, Y., and Mattina, M.
\newblock S2ta: Exploiting structured sparsity for energy-efficient mobile cnn acceleration.
\newblock In \emph{2022 IEEE International Symposium on High Performance Computer Architecture (HPCA)}, 2022{\natexlab{b}}.

\bibitem[Lu et~al.(2023)Lu, Agrawal, Subramanian, Rybakov, Sa, and Yazdanbakhsh]{lu2023step}
Lu, Y., Agrawal, S., Subramanian, S., Rybakov, O., Sa, C.~D., and Yazdanbakhsh, A.
\newblock {STEP: Learning N:M Structured Sparsity Masks from Scratch with Precondition}.
\newblock In \emph{ICML}, 2023.

\bibitem[Mahmoud et~al.(2020)Mahmoud, Edo, Zadeh, Mohamed~Awad, Pekhimenko, Albericio, and Moshovos]{mahmoud2020tensordash}
Mahmoud, M., Edo, I., Zadeh, A.~H., Mohamed~Awad, O., Pekhimenko, G., Albericio, J., and Moshovos, A.
\newblock Tensordash: Exploiting sparsity to accelerate deep neural network training.
\newblock In \emph{2020 53rd Annual IEEE/ACM International Symposium on Microarchitecture (MICRO)}, pp.\  781--795, 2020.
\newblock \doi{10.1109/MICRO50266.2020.00069}.

\bibitem[Mei et~al.(2021)Mei, Houshmand, Jain, Giraldo, and Verhelst]{zigzag}
Mei, L., Houshmand, P., Jain, V., Giraldo, S., and Verhelst, M.
\newblock Zigzag: Enlarging joint architecture-mapping design space exploration for dnn accelerators.
\newblock \emph{IEEE Transactions on Computers}, 70\penalty0 (8):\penalty0 1160--1174, 2021.
\newblock \doi{10.1109/TC.2021.3059962}.

\bibitem[Mishra et~al.(2021)Mishra, Latorre, Pool, Stosic, Stosic, Venkatesh, Yu, and Micikevicius]{mishra2021accelerating}
Mishra, A., Latorre, J.~A., Pool, J., Stosic, D., Stosic, D., Venkatesh, G., Yu, C., and Micikevicius, P.
\newblock Accelerating sparse deep neural networks.
\newblock \emph{arXiv preprint arXiv:2104.08378}, 2021.

\bibitem[Narang et~al.(2017)Narang, Undersander, and Diamos]{narang2017block}
Narang, S., Undersander, E., and Diamos, G.
\newblock Block-sparse recurrent neural networks.
\newblock \emph{arXiv preprint arXiv:1711.02782}, 2017.

\bibitem[Naumov et~al.(2019)Naumov, Mudigere, Shi, Huang, Sundaraman, Park, Wang, Gupta, Wu, Azzolini, et~al.]{dlrm}
Naumov, M., Mudigere, D., Shi, H.-J.~M., Huang, J., Sundaraman, N., Park, J., Wang, X., Gupta, U., Wu, C.-J., Azzolini, A.~G., et~al.
\newblock Deep learning recommendation model for personalization and recommendation systems.
\newblock \emph{arXiv preprint arXiv:1906.00091}, 2019.

\bibitem[Neuralmagic(2023)]{SparseZoo}
Neuralmagic.
\newblock Sparsezoo models, 2023.
\newblock \url{https://sparsezoo.neuralmagic.com/}.

\bibitem[NVIDIA(2020{\natexlab{a}})]{nvidia_ampere}
NVIDIA.
\newblock Nvidia ampere ga102 gpu architecture, 2020{\natexlab{a}}.
\newblock \url{https://www.nvidia.com/content/PDF/nvidia-ampere-ga-102-gpu-architecture-whitepaper-v2.1.pdf}.

\bibitem[NVIDIA(2020{\natexlab{b}})]{nvidia_volta}
NVIDIA.
\newblock Nvidia v100 tensor core gpu, 2020{\natexlab{b}}.
\newblock \url{https://images.nvidia.com/content/technologies/volta/pdf/volta-v100-datasheet-update-us-1165301-r5.pdf}.

\bibitem[NVIDIA(2021)]{nvidia_asp}
NVIDIA.
\newblock {NVIDIA ASP (Automatic Sparsity)}.
\newblock \url{https://github.com/NVIDIA/apex/tree/master/apex/contrib/sparsity}, 2021.

\bibitem[Parashar et~al.(2017)Parashar, Rhu, Mukkara, Puglielli, Venkatesan, Khailany, Emer, Keckler, and Dally]{parashar2017scnn}
Parashar, A., Rhu, M., Mukkara, A., Puglielli, A., Venkatesan, R., Khailany, B., Emer, J., Keckler, S.~W., and Dally, W.~J.
\newblock Scnn: An accelerator for compressed-sparse convolutional neural networks.
\newblock In \emph{2017 ACM/IEEE 44th Annual International Symposium on Computer Architecture (ISCA)}, pp.\  27--40, 2017.
\newblock \doi{10.1145/3079856.3080254}.

\bibitem[Parashar et~al.(2019)Parashar, Raina, Shao, Chen, Ying, Mukkara, Venkatesan, Khailany, Keckler, and Emer]{timeloop}
Parashar, A., Raina, P., Shao, Y.~S., Chen, Y.-H., Ying, V.~A., Mukkara, A., Venkatesan, R., Khailany, B., Keckler, S.~W., and Emer, J.
\newblock Timeloop: A systematic approach to dnn accelerator evaluation.
\newblock In \emph{2019 IEEE International Symposium on Performance Analysis of Systems and Software (ISPASS)}, pp.\  304--315, 2019.
\newblock \doi{10.1109/ISPASS.2019.00042}.

\bibitem[Paszke et~al.(2019)Paszke, Gross, Massa, Lerer, Bradbury, Chanan, Killeen, Lin, Gimelshein, Antiga, Desmaison, Kopf, Yang, DeVito, Raison, Tejani, Chilamkurthy, Steiner, Fang, Bai, and Chintala]{paszke2019pytorch}
Paszke, A., Gross, S., Massa, F., Lerer, A., Bradbury, J., Chanan, G., Killeen, T., Lin, Z., Gimelshein, N., Antiga, L., Desmaison, A., Kopf, A., Yang, E., DeVito, Z., Raison, M., Tejani, A., Chilamkurthy, S., Steiner, B., Fang, L., Bai, J., and Chintala, S.
\newblock Pytorch: An imperative style, high-performance deep learning library.
\newblock In Wallach, H., Larochelle, H., Beygelzimer, A., d\textquotesingle Alch\'{e}-Buc, F., Fox, E., and Garnett, R. (eds.), \emph{Advances in Neural Information Processing Systems 32}, pp.\  8024--8035. Curran Associates, Inc., 2019.

\bibitem[Pool \& Yu(2021)Pool and Yu]{pool2021channel}
Pool, J. and Yu, C.
\newblock Channel permutations for n:m sparsity.
\newblock In \emph{Advances in Neural Information Processing Systems}, volume~34, pp.\  13316--13327. Curran Associates, Inc., 2021.

\bibitem[PyTorch(2023)]{torchvision}
PyTorch.
\newblock Pytorch torchvision models, 2023.
\newblock \url{https://pytorch.org/vision/stable/index.html}.

\bibitem[Qin et~al.(2020)Qin, Samajdar, Kwon, Nadella, Srinivasan, Das, Kaul, and Krishna]{qin2020sigma}
Qin, E., Samajdar, A., Kwon, H., Nadella, V., Srinivasan, S., Das, D., Kaul, B., and Krishna, T.
\newblock Sigma: A sparse and irregular gemm accelerator with flexible interconnects for dnn training.
\newblock In \emph{2020 IEEE International Symposium on High Performance Computer Architecture (HPCA)}, pp.\  58--70, 2020.
\newblock \doi{10.1109/HPCA47549.2020.00015}.

\bibitem[Radford et~al.(2019)Radford, Wu, Child, Luan, Amodei, Sutskever, et~al.]{gpt2}
Radford, A., Wu, J., Child, R., Luan, D., Amodei, D., Sutskever, I., et~al.
\newblock Language models are unsupervised multitask learners.
\newblock \emph{OpenAI blog}, 1\penalty0 (8):\penalty0 9, 2019.

\bibitem[Ramachandran et~al.(2017)Ramachandran, Zoph, and Le]{ramachandran2017searching}
Ramachandran, P., Zoph, B., and Le, Q.~V.
\newblock Searching for activation functions.
\newblock \emph{arXiv preprint arXiv:1710.05941}, 2017.

\bibitem[Reddi et~al.(2020)Reddi, Cheng, Kanter, Mattson, Schmuelling, Wu, Anderson, Breughe, Charlebois, Chou, Chukka, Coleman, Davis, Deng, Diamos, Duke, Fick, Gardner, Hubara, Idgunji, Jablin, Jiao, John, Kanwar, Lee, Liao, Lokhmotov, Massa, Meng, Micikevicius, Osborne, Pekhimenko, Rajan, Sequeira, Sirasao, Sun, Tang, Thomson, Wei, Wu, Xu, Yamada, Yu, Yuan, Zhong, Zhang, and Zhou]{mlperf}
Reddi, V.~J., Cheng, C., Kanter, D., Mattson, P., Schmuelling, G., Wu, C.-J., Anderson, B., Breughe, M., Charlebois, M., Chou, W., Chukka, R., Coleman, C., Davis, S., Deng, P., Diamos, G., Duke, J., Fick, D., Gardner, J.~S., Hubara, I., Idgunji, S., Jablin, T.~B., Jiao, J., John, T.~S., Kanwar, P., Lee, D., Liao, J., Lokhmotov, A., Massa, F., Meng, P., Micikevicius, P., Osborne, C., Pekhimenko, G., Rajan, A. T.~R., Sequeira, D., Sirasao, A., Sun, F., Tang, H., Thomson, M., Wei, F., Wu, E., Xu, L., Yamada, K., Yu, B., Yuan, G., Zhong, A., Zhang, P., and Zhou, Y.
\newblock Mlperf inference benchmark.
\newblock In \emph{Proceedings of the ACM/IEEE 47th Annual International Symposium on Computer Architecture}, ISCA '20, pp.\  446–459. IEEE Press, 2020.
\newblock ISBN 9781728146614.
\newblock \doi{10.1109/ISCA45697.2020.00045}.

\bibitem[Rhu et~al.(2018)Rhu, O'Connor, Chatterjee, Pool, Kwon, and Keckler]{rhu2018cdma}
Rhu, M., O'Connor, M., Chatterjee, N., Pool, J., Kwon, Y., and Keckler, S.~W.
\newblock Compressing dma engine: Leveraging activation sparsity for training deep neural networks.
\newblock In \emph{2018 IEEE International Symposium on High Performance Computer Architecture (HPCA)}, pp.\  78--91, 2018.

\bibitem[Samajdar et~al.(2020)Samajdar, Joseph, Zhu, Whatmough, Mattina, and Krishna]{scale-sim}
Samajdar, A., Joseph, J.~M., Zhu, Y., Whatmough, P., Mattina, M., and Krishna, T.
\newblock A systematic methodology for characterizing scalability of dnn accelerators using scale-sim.
\newblock In \emph{2020 IEEE International Symposium on Performance Analysis of Systems and Software (ISPASS)}, pp.\  58--68, 2020.
\newblock \doi{10.1109/ISPASS48437.2020.00016}.

\bibitem[Shin et~al.(2021)Shin, Shafiee, Pedram, Abdel-Aziz, Li, and Hassoun]{shin2021griffin}
Shin, J.~H., Shafiee, A., Pedram, A., Abdel-Aziz, H., Li, L., and Hassoun, J.
\newblock Design space exploration of sparse accelerators for deep neural networks.
\newblock \emph{arXiv preprint arXiv:2107.12922}, 2021.

\bibitem[Shoeybi et~al.(2019)Shoeybi, Patwary, Puri, LeGresley, Casper, and Catanzaro]{shoeybi2019megatron}
Shoeybi, M., Patwary, M., Puri, R., LeGresley, P., Casper, J., and Catanzaro, B.
\newblock Megatron-lm: Training multi-billion parameter language models using model parallelism.
\newblock \emph{arXiv preprint arXiv:1909.08053}, 2019.

\bibitem[So et~al.(2021)So, Ma{\'n}ke, Liu, Dai, Shazeer, and Le]{so2021primer}
So, D.~R., Ma{\'n}ke, W., Liu, H., Dai, Z., Shazeer, N., and Le, Q.~V.
\newblock Primer: Searching for efficient transformers for language modeling.
\newblock \emph{arXiv preprint arXiv:2109.08668}, 2021.

\bibitem[Sun et~al.(2021)Sun, Zhou, Stuijk, Wijnhoven, Nelson, Li, and Corporaal]{sun2021domino}
Sun, W., Zhou, A., Stuijk, S., Wijnhoven, R., Nelson, A.~O., Li, h., and Corporaal, H.
\newblock Dominosearch: Find layer-wise fine-grained n:m sparse schemes from dense neural networks.
\newblock In Ranzato, M., Beygelzimer, A., Dauphin, Y., Liang, P., and Vaughan, J.~W. (eds.), \emph{Advances in Neural Information Processing Systems}, volume~34, pp.\  20721--20732. Curran Associates, Inc., 2021.

\bibitem[Thakker et~al.(2021)Thakker, Whatmough, Liu, Mattina, and Beu]{thakker2021doping}
Thakker, U., Whatmough, P., Liu, Z., Mattina, M., and Beu, J.
\newblock Doping: A technique for extreme compression of lstm models using sparse structured additive matrices.
\newblock \emph{Proceedings of machine learning and systems}, 3:\penalty0 533--549, 2021.

\bibitem[Wang et~al.(2021)Wang, Zhang, Xie, Guo, Liu, and Leng]{wang:dual-side}
Wang, Y., Zhang, C., Xie, Z., Guo, C., Liu, Y., and Leng, J.
\newblock Dual-side sparse tensor core.
\newblock In \emph{Proceedings of the 48th Annual International Symposium on Computer Architecture}, ISCA '21, pp.\  1083–1095. IEEE Press, 2021.
\newblock ISBN 9781450390866.
\newblock \doi{10.1109/ISCA52012.2021.00088}.

\bibitem[Wu et~al.(2021)Wu, Tsai, Parashar, Sze, and Emer]{wu2021sparseloop}
Wu, Y.~N., Tsai, P.-A., Parashar, A., Sze, V., and Emer, J.~S.
\newblock Sparseloop: An analytical, energy-focused design space exploration methodology for sparse tensor accelerators.
\newblock In \emph{2021 IEEE International Symposium on Performance Analysis of Systems and Software (ISPASS)}, pp.\  232--234, 2021.
\newblock \doi{10.1109/ISPASS51385.2021.00043}.

\bibitem[Wu et~al.(2022{\natexlab{a}})Wu, Tsai, Parashar, Sze, and Emer]{sparseloop-artifact}
Wu, Y.~N., Tsai, P.-A., Parashar, A., Sze, V., and Emer, J.~S.
\newblock {Artifact: Sparseloop: An Analytical Approach To Sparse Tensor Accelerator Modeling}.
\newblock Zenodo, oct 2022{\natexlab{a}}.
\newblock \doi{10.5281/zenodo.7027215}.

\bibitem[Wu et~al.(2022{\natexlab{b}})Wu, Tsai, Parashar, Sze, and Emer]{wu2022sparseloop}
Wu, Y.~N., Tsai, P.-A., Parashar, A., Sze, V., and Emer, J.~S.
\newblock Sparseloop: An analytical approach to sparse tensor accelerator modeling.
\newblock In \emph{2022 55th IEEE/ACM International Symposium on Microarchitecture (MICRO)}, pp.\  1377--1395, 2022{\natexlab{b}}.
\newblock \doi{10.1109/MICRO56248.2022.00096}.

\bibitem[Wu et~al.(2023)Wu, Tsai, Muralidharan, Parashar, Sze, and Emer]{highlight2023micro}
Wu, Y.~N., Tsai, P.-A., Muralidharan, S., Parashar, A., Sze, V., and Emer, J.
\newblock Highlight: Efficient and flexible dnn acceleration with hierarchical structured sparsity.
\newblock In \emph{Proceedings of the 56th Annual IEEE/ACM International Symposium on Microarchitecture}, MICRO '23, pp.\  1106–1120, New York, NY, USA, 2023. Association for Computing Machinery.
\newblock ISBN 9798400703294.
\newblock \doi{10.1145/3613424.3623786}.

\bibitem[Xiao(2023)]{Moffett_antoum}
Xiao, Z.
\newblock Moffett antoum®: A deep-sparse ai inference system-on-chip for vision and large-language models.
\newblock In \emph{2023 IEEE Hot Chips 35 Symposium (HCS)}, pp.\  1--33, 2023.
\newblock \doi{10.1109/HCS59251.2023.10254723}.

\bibitem[Yazdanbakhsh et~al.(2022)Yazdanbakhsh, Kao, Agrawal, Subramanian, Krishna, and Evci]{NM_training_recipe}
Yazdanbakhsh, A., Kao, S.-C., Agrawal, S., Subramanian, S., Krishna, T., and Evci, U.
\newblock Training recipe for n:m structured sparsity with decaying pruning mask, 2022.

\bibitem[Ye et~al.(2023)Ye, Lai, Shao, Chen, and Ceze]{ye2023sparsetir}
Ye, Z., Lai, R., Shao, J., Chen, T., and Ceze, L.
\newblock Sparsetir: Composable abstractions for sparse compilation in deep learning.
\newblock In \emph{Proceedings of the 28th ACM International Conference on Architectural Support for Programming Languages and Operating Systems, Volume 3}, ASPLOS 2023, pp.\  660–678, New York, NY, USA, 2023.
\newblock \doi{10.1145/3582016.3582047}.

\bibitem[Zaheer et~al.(2020)Zaheer, Guruganesh, Dubey, Ainslie, Alberti, Ontanon, Pham, Ravula, Wang, Yang, et~al.]{zaheer2020big}
Zaheer, M., Guruganesh, G., Dubey, K.~A., Ainslie, J., Alberti, C., Ontanon, S., Pham, P., Ravula, A., Wang, Q., Yang, L., et~al.
\newblock Big bird: Transformers for longer sequences.
\newblock \emph{Advances in neural information processing systems}, 33:\penalty0 17283--17297, 2020.

\bibitem[Zhang et~al.(2022)Zhang, Lin, Lin, Luo, Li, Chao, Wu, and Ji]{zhang2022learning}
Zhang, Y., Lin, M., Lin, Z., Luo, Y., Li, K., Chao, F., Wu, Y., and Ji, R.
\newblock Learning best combination for efficient n: M sparsity.
\newblock \emph{Advances in Neural Information Processing Systems}, 35:\penalty0 941--953, 2022.

\bibitem[Zhou et~al.(2021)Zhou, Ma, Zhu, Liu, Zhang, Yuan, Sun, and Li]{zhou2021nm}
Zhou, A., Ma, Y., Zhu, J., Liu, J., Zhang, Z., Yuan, K., Sun, W., and Li, H.
\newblock Learning n:m fine-grained structured sparse neural networks from scratch.
\newblock In \emph{International Conference on Learning Representations}, 2021.

\bibitem[Zhu et~al.(2019)Zhu, Zhang, Gu, and Xie]{zhu2019micro}
Zhu, M., Zhang, T., Gu, Z., and Xie, Y.
\newblock Sparse tensor core: Algorithm and hardware co-design for vector-wise sparse neural networks on modern gpus.
\newblock In \emph{Proceedings of the 52nd Annual IEEE/ACM International Symposium on Microarchitecture}, MICRO '52, pp.\  359–371, New York, NY, USA, 2019. Association for Computing Machinery.
\newblock ISBN 9781450369381.

\end{thebibliography}
\bibliographystyle{mlsys2025}

\appendix
\newpage
\section{Additional details of TASD}
\label{append:additional_details}
\subsection{An analysis of TASD with synthetic data}
\label{sec:tasd_analysis}

\begin{figure}[]
	\centering
	\includegraphics[width=0.47\textwidth]{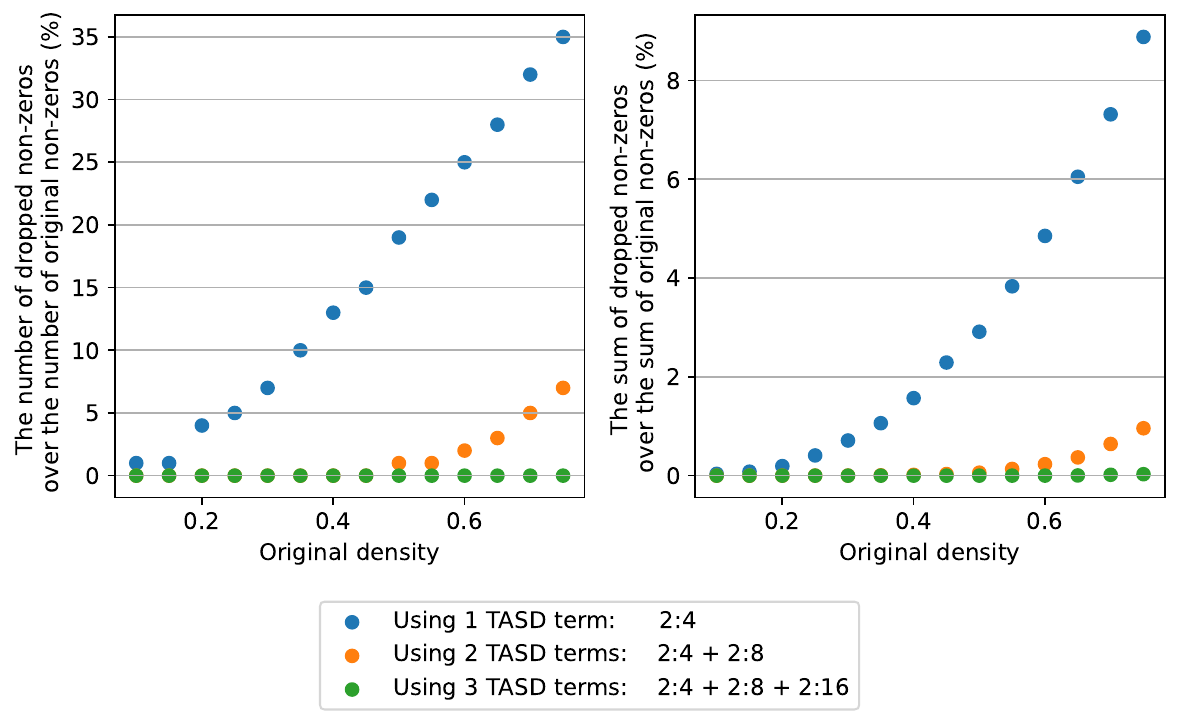}
	\caption{Percentages of dropped non-zeros and the sum of dropped non-zeros after applying different TASD series.}
	\label{fig:normal_result}
\end{figure}

The number of dropped non-zeros and the sum of the dropped magnitudes are crucial as they correlate to the potential loss of accuracy when applying TASD.
Thus, we first conduct preliminary experiments with synthetic data using various TASD series and matrices to understand the trade-offs.

We generate a synthetic matrix with dimensions of 128$\times$128 and densities ranging from 0.1 to 0.75.
We explore three TASD series in this experiment; 1) using one term with 2:4, 2) using two terms with 2:4 and 2:8, and 3) using three terms with 2:4, 2:8, and 2:16.
To consider various distributions, we tested two different distributions, a uniform random distribution between 0 and 1 and a normal distribution with a mean of 0 and a standard deviation of $\frac{1}{3}$.
\autoref{fig:normal_result} shows the results with matrices generated using the normal distribution.

\textbf{Takeaways}: 
1) If the matrix is very sparse, the percentage of dropped non-zero values becomes noticeably small, less than 1\%, even with just two TASD terms.
2) Since we choose elements with a greedy approach (i.e., keep the largest non-zero), the percentage of dropped total magnitude is lower than the percentage of dropped non-zero values, allowing better approximation even for higher densities.

In addition, we also find that across different distributions, percentages of dropped non-zero values are similar since they depend on the density of the original matrix, but percentages of the dropped total magnitude vary slightly. Interestingly, we observed that Mean Square Errors (MSEs) vary significantly depending on the distribution. This implies that not the sparsity degree only, but the actual distribution is also critical for finding a high-quality TASD series configuration.

\begin{figure}[t]
	\centering
	\includegraphics[width=0.49\textwidth]{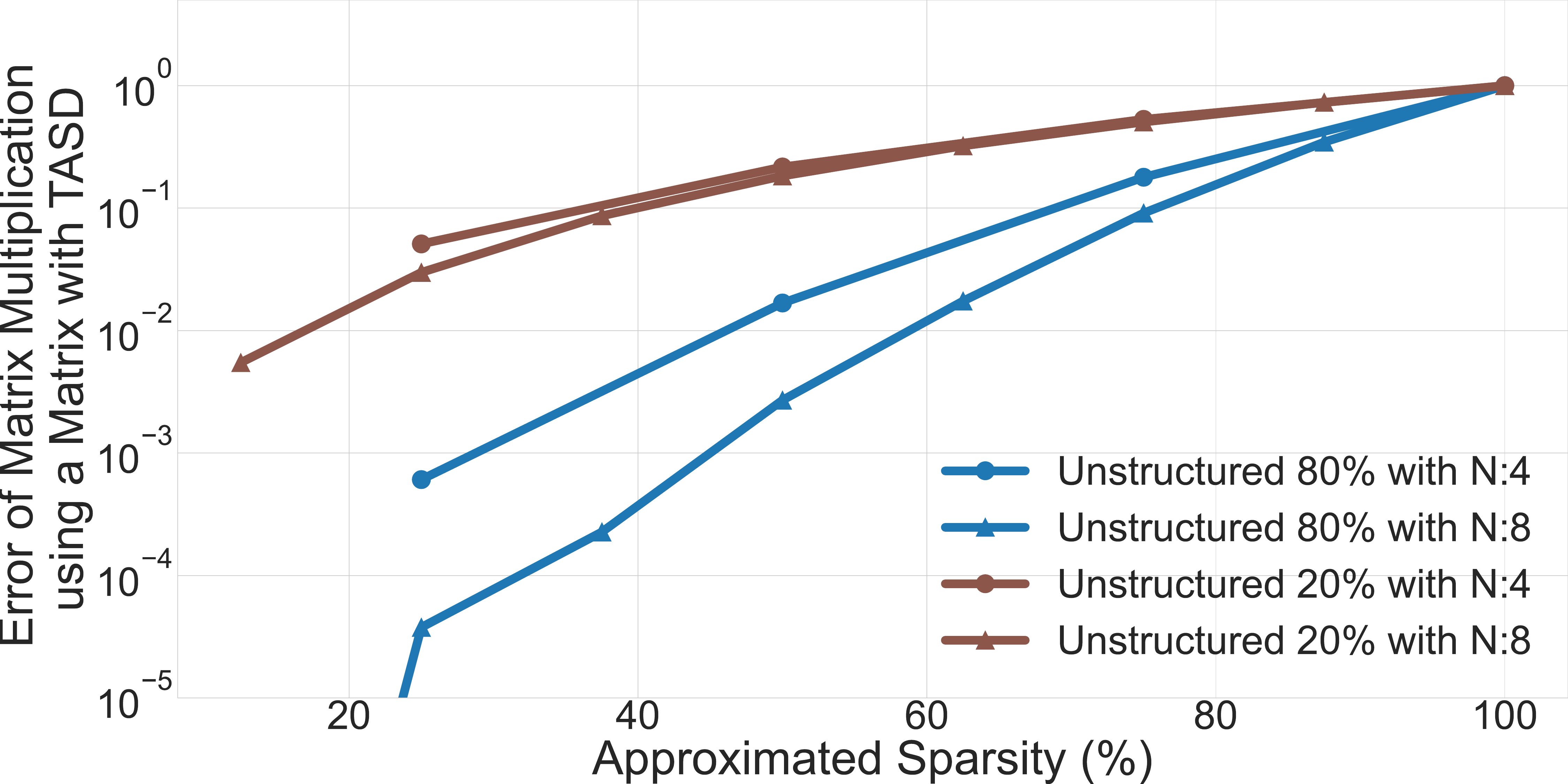}
	\caption{The error between the result from the original matrices and the result with an approximated matrix using TASD assuming different sparsity in the original matrix.
        }
	\label{fig:matmul_error}
\end{figure}

\textbf{Using TASD for Matrix Multiplication: }
To understand the impact of using TASD for matrix multiplication, we run another experiment using matrices $A$ and $B$ with the dimensions of 256$\times$256. 
We set each element to have a random value between 0 and 1. 
For matrix $A$, we generate unstructured sparsity with two sparsity degrees 20\% and 80\%, and we keep $B$ as dense.
Then, we apply one-term TASD on $A$ with 0-4:4 and 0-8:8 TASD configurations.
We measure the error as the Frobenius norm of the result with approximated operands divided by the original Frobenius norm, $\frac{||(A-A^*)B||}{||AB||}$.
We represent configurations with \emph{approximated sparsity}, which means the sparsity degree of a structured sparse pattern. For example, 1:4 pattern and 2:8 pattern both have an approximated sparsity of 75\%.
We plot the errors with different TASD configurations and approximated sparsity degrees in \autoref{fig:matmul_error}.

\textbf{Error Behavior:} The first trend we observe is that the error gets smaller as the approximated sparsity gets lower since it is likely to drop fewer non-zeros with a more conservative approximation.
Second, for the same approximated sparsity and the block size (M), the error gets smaller as the matrix $A$ gets sparser. 
Given the same TASD configuration, the sparser matrices would drop fewer non-zeros using the same TASD configuration as shown in \autoref{fig:normal_result}, thus resulting in a smaller error.
Third, with the same sparsity of matrix $A$ and approximated sparsity, the N:4 configuration causes a larger error than the N:8 configuration (such as 1:4 and 2:8), since the expressiveness of the N:8 pattern is higher.
Finally, given any unstructured sparse tensor, \emph{we can limit the error of matrix multiplication by conservatively selecting the TASD configuration, while maximizing the compute reduction}.
This optimization thus becomes the key to leveraging TASD for accelerating sparse DNN models with structured sparse hardware. In terms of the number of effectual computations, using a sparser TASD configuration is favorable to approximate the matrix multiplication while it would cause a higher error if not chosen cautiously.

\section{Additional Experiments}
\label{append:additional_experiments}
In this section, we provide additional experiments to understand the impact of TASD.
\subsection{
Comparison to structured sparse accelerators
}
\begin{figure}[!t]
	\centering
	\includegraphics[width=0.49\textwidth]{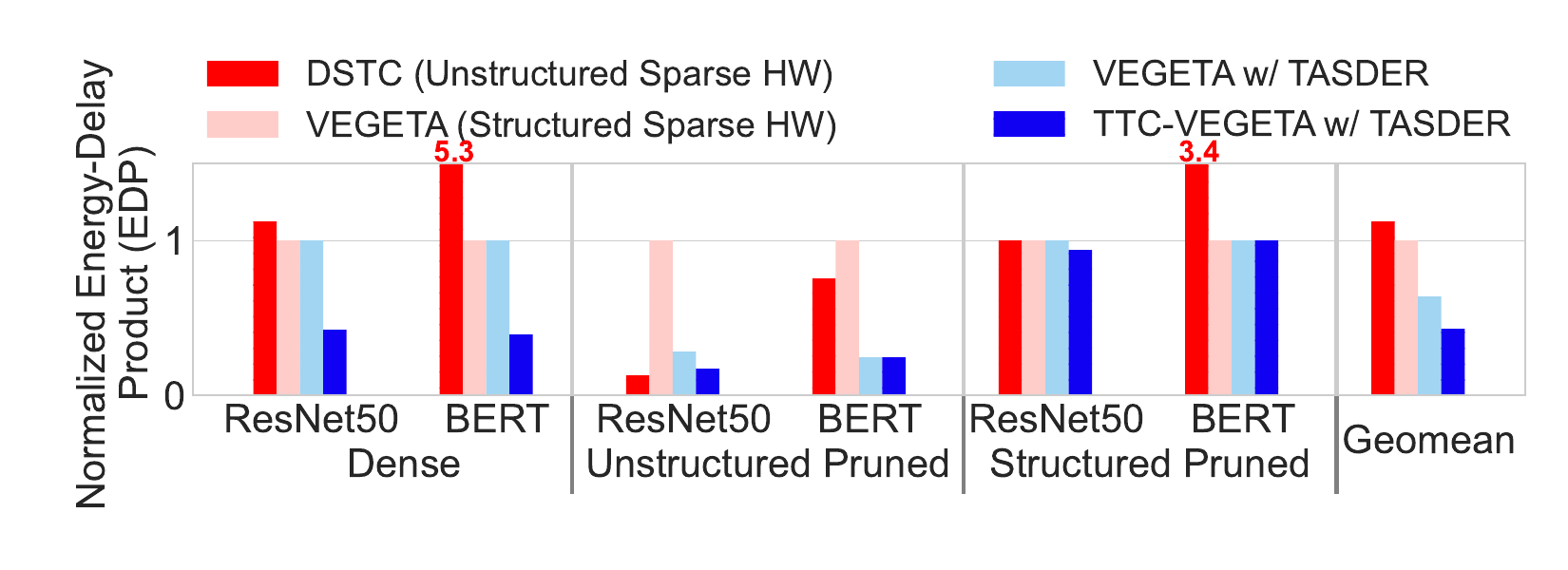}
        \vspace{-1em}

	\caption{
        Study on DSTC, VEGETA, TASDER, and TTC with different types of models.
 }
	\label{fig:vegeta-comparison}
 \end{figure}
To study how the proposed TTC-based accelerator compared to prior structured sparsity accelerators, we conduct an ablation study to show how different novelties in this work contribute to the efficiency gain.
\autoref{fig:vegeta-comparison} shows the normalized EDP improvement for four different system: DSTC, VEGETA without TASDER, VEGETA with TASDER, TTC-VEGETA with TASDER.
Without TASDER and HW-aware fine-tuning, VEGETA cannot exploit sparsity in off-the-shelf DNNs and has no improvement at all.
If the model is structured pruned using HW-aware fine-tuning, VEGETA can exploit sparsity achieving a comparable EDP to TTC-VEGETA.
With TASDER, VEGETA can exploit weight sparsity in unstructured sparse ResNet50/BERT since TASDER transforms unstructured sparse weights into structured sparsity supported by VEGETA.
Finally, with dynamic decomposition support for activation sparsity, TTC-VEGETA can also exploit activation sparsity, further improving EDP for all DNNs.

\subsection{TASD on more DNN models}
\begin{figure}[!t]
	\centering
	\includegraphics[width=0.5\textwidth]{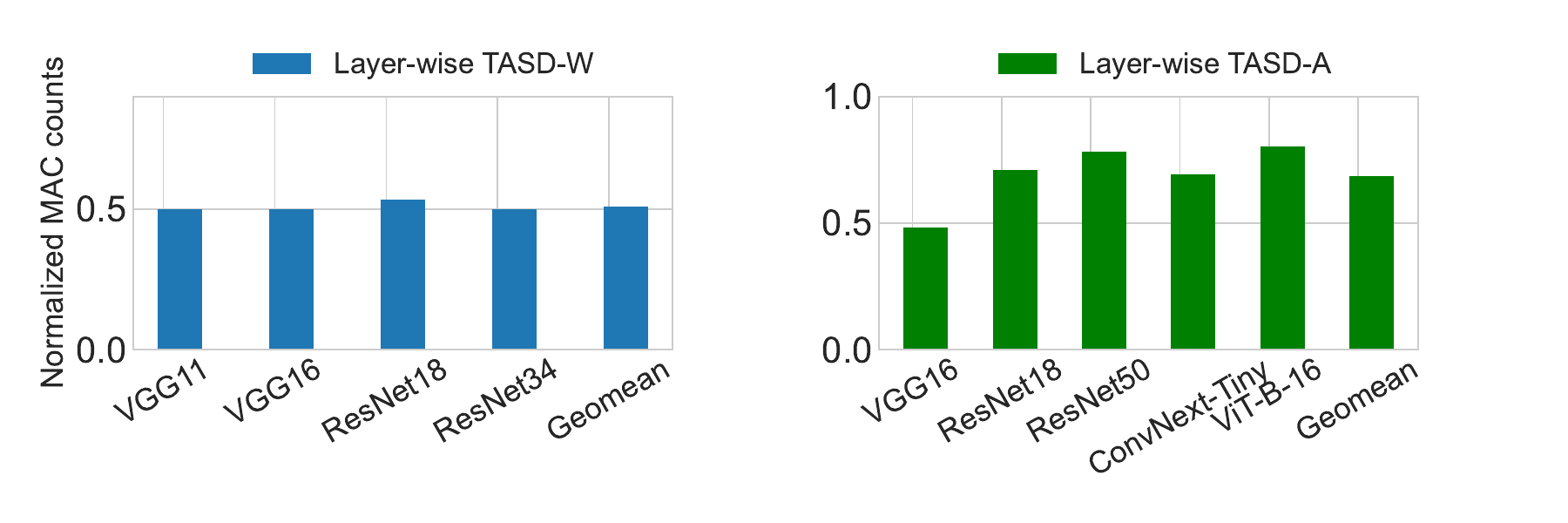}
    \vspace{-1em}
	\caption{Layer-wise TASD on more DNN models. 
 }
	\label{fig:maccounts-tasd-w}
\end{figure}

To further investigate the impact of TASD-W on different sparse DNNs, we applied layer-wise TASD-W on different Sparse ResNet and VGG families with a requirement to maintain 99\% of the original accuracy. 
We use the pre-trained unstructured sparse models from SparseZoo~\cite{SparseZoo}.
Across different sparse ResNet models and VGG models, TASD-W reduced 49\% MAC operations while maintaining 99\% accuracy, as shown in \autoref{fig:maccounts-tasd-w}.

On the other hand, to understand the potential impact of TASD-A on other DNN models, we applied TASD-A on various models including both convolution networks and a transformer-based network, as shown in \autoref{fig:maccounts-tasd-w}.
We use the pre-trained dense models from TorchVision~\cite{torchvision} and Huggingface for this evaluation and we use the requirement to meet 99\% of the original accuracy. 
We observe that the layerwise TASD-A is effective for various models and achieves 32\% reduction in MACs for other models on average.

\end{document}